%% file: main.tex
\documentclass[conference]{IEEEtran}
\IEEEoverridecommandlockouts

\usepackage{amsmath,amssymb,amsfonts}
\usepackage{algorithmic}
\usepackage{graphicx}
\usepackage{textcomp}
\usepackage{xcolor}
\usepackage[numbers]{natbib}
\usepackage[utf8]{inputenc} 
\usepackage[T1]{fontenc}    
\usepackage{hyperref}       
\usepackage{url}            
\usepackage{booktabs}       
\usepackage{nicefrac}       
\usepackage{microtype}      
\usepackage{color}
\usepackage{wrapfig}
\usepackage{enumerate}
\usepackage{bm}
\usepackage{algorithm}
\usepackage{array}
\usepackage{scalefnt}
\usepackage{makecell}
\usepackage{mathrsfs}
\usepackage{flushend, multirow}
\usepackage{lipsum}
\usepackage{enumitem,calc}

\newcommand{\eg}{{\sl e.g.}}
\newcommand{\ie}{{\sl i.e.}}

\newcommand{\method}{OCEAN}

\newcommand{\nop}[1]{}
\newcommand{\lc}[1]{{\textsf{\textcolor{green!10!orange!90!}{[From LC: #1]}}}}
\newcommand{\zc}[1]{{\textsf{\textcolor{blue}{[From Zach: #1]}}}}

\def\BibTeX{{\rm B\kern-.05em{\sc i\kern-.025em b}\kern-.08em
    T\kern-.1667em\lower.7ex\hbox{E}\kern-.125emX}}
\begin{document}

\newcommand{\mkclean}{
  \renewcommand{\zc}[1]{}
  \renewcommand{\lc}[1]{}
}

\makeatletter
\def\@makefntext#1{\noindent\makebox[0pt][r]{\@thefnmark.\,}#1}
\renewcommand{\footnoterule}{
  \vspace*{0.8em}
  \hrule width 0.5\columnwidth height 0.4pt
  \vspace*{0.6em}
}
\makeatother


\title{Online Multi-modal Root Cause Identification in Microservice Systems}


\author{Lecheng Zheng$^{1, 2}$, Zhengzhang Chen$^{2,*}$, Haifeng Chen$^{2}$\\
$^{1}$University of Illinois Urbana-Champagin, $^{2}$NEC Labs America\\
\texttt{lecheng4@illinois.edu, zchen@nec-labs.com, haifeng@nec-labs.com}
\thanks{$^{*}$Corresponding author}
}

\maketitle

\input{Abstract}


\input{Introduction}

\input{Related_work}
\input{Methodology}
\input{Experiment}
\input{Conclusion}
\bibliographystyle{IEEEtran}
\bibliography{short_reference}
\input{Appendix}

\end{document}

%% file: Abstract.tex
\begin{abstract}
Root Cause Analysis (RCA) is essential for pinpointing the root causes of failures in microservice systems. Traditional data-driven RCA methods are typically limited to offline applications due to high computational demands, and existing online RCA methods handle only single-modal data, overlooking complex interactions in multi-modal systems. In this paper, we introduce \method, a novel online multi-modal causal structure learning method for root cause localization. \method\ introduces a long-term temporal causal learning module with two encoders: one captures stable causal dependencies from historical data, while the other models short-term variations in the current batch data. We further design a multi-factor attention mechanism to analyze and reassess the relationships among different metrics and log indicators/attributes for enhanced online causal graph learning. Additionally, a contrastive mutual information maximization-based graph fusion module is developed to effectively model the relationships across various modalities. Extensive experiments on three real-world datasets demonstrate the effectiveness and efficiency of our proposed method.
\end{abstract}

%% file: Introduction.tex
\section{Introduction}
\label{sec:intro}
Root Cause Analysis (RCA) is crucial for identifying the underlying causes of system failures and ensuring the high performance of microservice systems~\citep{wang2023incremental, li2021practical}. Traditional manual root cause analysis is labor-intensive, costly, and error-prone, given the complexity of microservice systems and the extensive volume of data involved. Consequently, effective and efficient root cause analysis methods are vital for pinpointing failures in complex microservice systems and mitigating potential financial losses when system faults occur. Prior works leveraging causal discovery have focused on constructing causal or dependency graphs~\citep{DBLP:conf/nips/IkramCMSBK22, lu2017log, li2021practical}, which reveal causal links among system entities and key performance indicators to trace underlying faults.

Despite significant advances, most of these approaches are designed for offline use and face challenges with real-time implementation in microservice systems due to high computational demands. To address this, Wang \textit{et al.}~\citep{wang2023incremental} introduced an online RCA method that decouples state-invariant and state-dependent information and incrementally updates the causal graph. Li \textit{et al.}~\citep{DBLP:conf/kdd/0005LYNZSP22} developed a causal Bayesian network that leverages system architecture knowledge to mitigate potential biases toward new data. However, these online RCA methods are limited to handling single-modal data. Recently, multi-modal data, such as system metrics and logs, are commonly collected from microservice systems, revealing the complex nature of system failures~\citep{DBLP:conf/www/ZhengCHC24}. For instance, failures such as ``Database Query Failures'' might be overlooked if only system metrics are considered, whereas issues like ``Disk Space Full'' are more effectively identified through combined analysis of metrics and logs. This underscores the importance of using multi-modal data for a thorough understanding of system failures. By integrating information from various sources, we can detect the abnormal patterns of system failures that is evident when analyzing single-modal data.

To bridge this gap, this paper proposes an online multi-modal causal structure learning method for RCA in microservice systems. Given the system KPI and multi-modal data (metrics and log data), our objective is to construct an online multi-modal causal graph that identifies the top $k$ system entities most relevant to the system KPI. Three major challenges arise in this task: (C1) \textbf{Enabling Long-term Temporal Causal Relationship Learning in the Online Setting}: Existing temporal/sequential modeling techniques~\citep{chen2024automatic, shan2024face, zhou2024causalkgpt, wang2023incremental,  DBLP:conf/www/ZhengCHC24}, (e.g., RNNs, Transformers, and LLMs) are computationally expensive and limited to short-term dependencies. 
However, some system faults, such as Distributed Denial of Service (DDoS) attacks, may persist for extended periods. Effectively capturing these long-term temporal dependencies in the online setting is crucial for efficiently identifying various types of system faults. (C2) \textbf{Capturing the Correlation of Multi-dimensional Factors}: Existing RCA approaches~\citep{DBLP:conf/nips/IkramCMSBK22, wang2023interdependent, DBLP:conf/www/ZhengCHC24} often analyze abnormal patterns from multiple factors individually, such as CPU usage or memory usage from system metrics and frequency or golden signal from system logs, overlooking potential relationships among these factors from both modalities. Furthermore, these methods often consider all factors as equally important; however, in real applications, certain factors prove to be considerably more crucial than others. It is vital, therefore, to reassess the contributions of each factor to the learning of causal structures.
(C3) \textbf{Learning Multi-modal Causal Structures}: Effectively capturing the relationships between different modalities~\cite{DBLP:conf/www/ZhouZZLH20, DBLP:conf/acl/0006ZJFJBH025, DBLP:journals/corr/abs-2502-08942, DBLP:journals/corr/abs-2504-07394, DBLP:conf/kdd/ZhengJLTH24, DBLP:journals/corr/abs-2102-07751, DBLP:conf/sdm/ZhengCH19} in an online setting is crucial. Simply combining causal graphs from individual modalities can be problematic, especially if one modality is of lower quality.

To tackle these challenges, we introduce \method, \underline{O}nline Multi-modal \underline{C}ausal Structure L\underline{EA}r\underline{N}ing, for root cause identification in microservice systems. Specifically, we design a long-term temporal causal relationship learning module with two encoders, one for capturing the long-term temporal dependencies and invariant causal relation for historical data and another for modeling the small changes within current batch data. We further develop a multi-factor attention mechanism to analyze the correlations among various factors and reassess their importance for causal graph learning. Additionally, we propose a contrastive mutual information estimation technique to model the relationships of different modalities. Our contributions can be summarized as follows:

\begin{itemize}
    \item We introduce a novel online framework for multi-modality root cause analysis.
    \item We propose long-term temporal causal relationship learning module with two encoders, aiming to efficiently capture long-term temporal dependencies and causal relations for both historical data and current batch data.
    \item We develop graph fusion techniques with contrastive multi-modal learning to model the relationships between different modalities and assess their importance.
    \item Extensive experiments on three real-world datasets demonstrate the effectiveness and efficiency of our proposed method.
\end{itemize}

%% file: Related_work.tex
\section{Preliminary and Related Work}
\label{sec:pre}
\noindent\textbf{Key Performance Indicator (KPI)}. In a microservice system, KPIs serve as invaluable metrics for assessing the effectiveness and productivity of the architecture~\citep{podgorski2015measuring}. They play an indispensable role in monitoring and managing different aspects of microservices to uphold optimal performance levels. Common KPIs encompass latency and service response time. High values in these metrics typically indicate suboptimal performance or potential failure.

\noindent\textbf{Entity Metrics}. 
Entity metrics are the measurable time-series attributes that provide insights into the performance and status of services within a system~\citep{bogner2017automatically}.
These entities encompass various components such as physical machines, containers, virtual machines, and pods. In microservice architectures, typical entity metrics include CPU utilization, memory usage, disk I/O activity, packet transmission rate, and etc. These metrics are extensively employed to detect anomalous behavior and pinpoint potential causes of system failures in microservice environments~\citep{DBLP:conf/www/ZhengCHC24, soldani2022anomaly}.

\noindent\textbf{Root Cause Analysis}.
Current root cause analysis (RCA) methods can be categorized into two main branches: single-modal RCA methods and multi-modal RCA methods. Single-modal RCA methods primarily investigate causal relationships among system components using one type of data only~\citep{soldani2022anomaly, li2021practical,wang2023interdependent,wang2023incremental}, while multi-modal RCA methods~\citep{yu2023nezha, DBLP:conf/ispa/HouJWLH21, DBLP:conf/www/ZhengCHC24} benefit from leveraging the rich data sources to achieve better performance. Recently, large language model (LLM)-based approaches have emerged as a new research direction for learning causal relations in root cause identification, owing to the success of LLMs in tackling complex tasks~\citep{chen2024automatic, shan2024face, goel2024x, zhou2024causalkgpt, roy2024exploring, wang2023rcagent}. Unlike existing RCA methods, this paper addresses the online multi-modal RCA problem by uniquely modeling long-term temporal dependencies while simultaneously capturing the cross-modal correlation of multiple factors.  

%% file: Methodology.tex
\section{Methodology}
\begin{figure*}[!t]
\centering
\vspace{-5pt}
\includegraphics[width=0.95\linewidth]{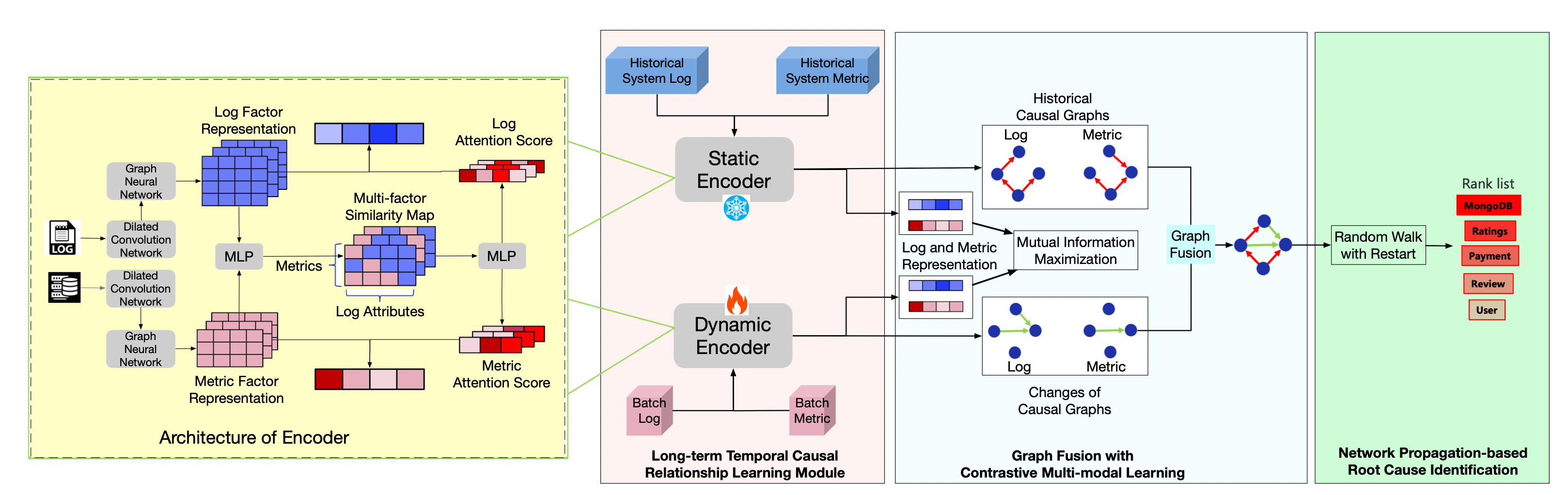} \\
\caption{
The overview of the proposed framework \method\ with three main modules: long-term temporal causal relationship learning, graph fusion with contrastive multi-modal learning, and network propagation-based root cause localization. Both static encoder and dynamic encoder are trained in the offline stage but only dynamic encoder will be finetuned with batch (streaming) data during the online stage.} 
\label{fig:ocean_architecture}
\vspace{-5pt}
\end{figure*}

In this section, we first present the problem statement and then introduce \method, an online causal structural learning method designed to identify root causes using multi-modal data. We propose three modules to tackle the challenges outlined in the introduction: long-term temporal causal relationship learning, contrastive multi-modal learning and network propagation-based root cause identification module. The overview of \method\ is provided in Figure~\ref{fig:ocean_architecture}.

\subsection{Problem Statement}

Let $\mathcal{X}_M = \{\mathbf{X}_M^0, \mathbf{X}_M^1, \ldots, \mathbf{X}_M^{\mathcal{T}}\}$ denote the multivariate time-series data of \textbf{system metrics} across $n-1$ entities, where $\mathbf{X}_M^0 \in \mathbb{R}^{(n-1) \times d_M \times T_1}$ represents a large historical dataset and $\mathbf{X}_M^i \in \mathbb{R}^{(n-1) \times d_M \times T_2}$ for $i \in [1, \mathcal{T}]$ are smaller sequential streaming batches. 
Here, $d_M$ is the number of metric features, and $T_1$ and $T_2$ denote the temporal lengths of the historical and streaming segments, respectively. 
Similarly, $\mathcal{X}_L = \{\mathbf{X}_L^0, \mathbf{X}_L^1, \ldots, \mathbf{X}_L^{\mathcal{T}}\}$
denotes the multivariate time-series data of \textbf{system logs}, where each 
$\mathbf{X}_L^i \in \mathbb{R}^{(n-1) \times d_L \times T}$ is obtained by preprocessing log records into structured time-series format, and $d_L$ is the number of log features. The \textbf{system KPI} is represented as $\mathbf{y} = \{\mathbf{y}^0, \mathbf{y}^1, \ldots, \mathbf{y}^{\mathcal{T}}\},$
where $\mathbf{y}^0 \in \mathbb{R}^{T_1}$ and $\mathbf{y}^i \in \mathbb{R}^{T_2}$ correspond to historical and batch KPI data, respectively. We consider $n$ nodes in total: $n-1$ microservice entities and one KPI node. The causal structure among them is modeled by a directed graph $\mathcal{G} = (\mathcal{V}, \mathcal{A}),$ where $\mathcal{V}$ denotes the set of vertices and $\mathcal{A} \in \mathbb{R}^{n \times n}$ is the adjacency matrix, with $\mathcal{A}_{ij} \neq 0$ indicating a causal influence from node $i$ to node $j$.
For unified processing, we replicate the KPI $d_M$ times and concatenate it with the metric and log data, forming $\hat{\mathbf{X}}_M^0 \in \mathbb{R}^{n \times d_M \times T_1}, \quad \hat{\mathbf{X}}_M^i \in \mathbb{R}^{n \times d_M \times T_2},$ and $\hat{\mathbf{X}}_L^0 \in \mathbb{R}^{n \times d_L \times T_1}, \quad \hat{\mathbf{X}}_L^i \in \mathbb{R}^{n \times d_L \times T_2}.$ In the \textbf{offline setting}, only the historical data ($\mathbf{X}^0$, $\mathbf{y}^0$) are available for model training, while in the \textbf{online setting}, streaming batches ($\mathbf{X}^i$, $\mathbf{y}^i$) are also used for root cause analysis. Finally, we employ the \textbf{Multivariate Singular Spectrum Analysis (MSSA)} model~\citep{alanqary2021change}, a state-of-the-art method for online anomaly detection, to trigger the root cause analysis process. 
We summarize the notations in Table~\ref{table_notation}.

\begin{table}
\caption{Notation Table}
\vspace{-3mm}
\centering
\scalebox{0.95}{
\begin{tabular}{*{2}{c}}
\hline    
$\bm{X}_M^{0}$  & the historical system metric (offline data) \\
$\bm{X}_M^{i}$  & the $i$-th batch of the system metric (streaming data)\\
$\bm{X}_L^{0}$  & the historical system log (offline data)\\
$\bm{X}_L^{i}$  & the $i$-th batch of system log (streaming data) \\
$T_1$ & the length of the historical metric data \\
$T_2$ & the length of the batch for the system metric \\
$n-1$ & the number of system entities \\ 
$\mathcal{T}$   & the total number of batches \\
$d_M$ & the number of different system metric features \\
$d_L$ & the number of different system log features \\
$\mathbf{y}$ & the system Key Performance Indicator \\
$\mathcal{G}=\{\mathcal{V}, \mathcal{A}\}$ & the causal graph \\
$\mathcal{A} $ & the adjacency matrix in the causal graph \\
\hline
\end{tabular}}
\label{table_notation}
\vspace{-5mm}
\end{table} 

\subsection{Long-term Temporal Causal Relationship Learning}
\label{long_term_TCS}
To effectively capture temporal causal relationships among system entities and KPIs, we adopt the Vector Autoregression (VAR) model~\citep{stock2001vector}, a classical yet powerful framework for modeling dynamic interactions in multivariate time series.  
Given the two-dimensional matrix $\mathbf{X}_{M,i} \in \mathbb{R}^{n \times T_1}$ representing the $i^{\textrm{th}}$ system metric, the model predicts future values as:
\begin{align}
    \nonumber \hat{\mathbf{X}}_{M,i}^{t} &= \mathbf{W}^1 \mathbf{X}_{M,i}^{t-1} + \mathbf{W}^2 \mathbf{X}_{M,i}^{t-2} + \cdots + \mathbf{W}^{t-1} \mathbf{X}_{M,i}^{1} + \boldsymbol{\epsilon}, \\
    \mathcal{L}_{\text{var}} &= \sum_{i=1}^{d_M} \|\mathbf{X}_{M,i}^{t} - F(\hat{\mathbf{X}}_{M,i}^{t}, \mathcal{A}, \theta)\|^2,
\end{align}
where $\hat{\mathbf{X}}_{M,i}^{t}$ denotes the predicted values for the $i^{\textrm{th}}$ metric, $\mathbf{W}^k \in \mathbb{R}^{n \times n}$ are lag-specific weight matrices, $\boldsymbol{\epsilon}$ is the residual error, and $F(\cdot)$ represents a graph neural network (GNN)~\citep{DBLP:conf/iclr/KipfW17} parameterized by $\theta$.  
The learnable adjacency matrix $\mathcal{A}$ encodes causal relationships among entities and the KPI.  
Although a $t^{\textrm{th}}$-order VAR model captures long-range dependencies, it becomes computationally expensive as $t$ increases~\citep{lin2020limitations}. Similarly, recurrent architectures (e.g., RNNs) and Transformers~\citep{DBLP:conf/nips/VaswaniSPUJGKP17} often suffer from high computational overhead, limiting their suitability for online root cause analysis.
To address these challenges, we propose a dual-encoder framework consisting of a \textit{Static Encoder} and a \textit{Dynamic Encoder} to efficiently learn long-term temporal dependencies and causal structures for \textit{historical} and \textit{streaming} data, respectively.  
The Static Encoder extracts stable, invariant causal relationships among entities in the offline stage, while the Dynamic Encoder incrementally adapts the learned causal structure during online updates.

\paragraph{Static Encoder}
In microservice systems, each entity is associated with multiple metric and log indicators (e.g., CPU usage, memory usage, log frequency, golden signals, etc.).  Conventional RCA methods typically analyze each factor independently, overlooking cross-factor dependencies and their varying importance under different abnormal patterns.  To address this limitation, we design the Static Encoder to jointly model temporal dependencies and multi-factor correlations across two modalities (metrics and logs) using attention-based multi-factor learning~\citep{DBLP:conf/nips/LuYBP16, DBLP:conf/nips/VaswaniSPUJGKP17}.
Given historical metric data $\hat{\mathbf{X}}_M^0$ and log data $\hat{\mathbf{X}}_L^0$, we first capture temporal dependencies via a gated temporal convolutional network (TCN)~\citep{DBLP:conf/ijcai/WuPLJZ19}:
\begin{align}
\label{TCN}
    g(\mathbf{x}, \mathbf{f}) &= \mathbf{x} * \mathbf{f} = \sum_{\tau=0}^{K-1} \mathbf{f}(\tau) \cdot \mathbf{x}(t - d \times \tau), \\
    \label{temp_s0_historical}
    \mathbf{H}_v^0 &= \tanh(g(\hat{\mathbf{X}}_v^0, \mathbf{f}_1)) \odot \sigma(g(\hat{\mathbf{X}}_v^0, \mathbf{f}_2)),
\end{align}
where $\mathbf{f}\in\mathbb{R}^K$ represents the 1-D kernel,  $\mathbf{f}_1, \mathbf{f}_2\in\mathbb{R}^K$ are 1-D convolution kernels, $d$ is the dilation factor, $\odot$ denotes the Hadamard product, $\sigma(\cdot)$ and $\tanh(\cdot)$ are the sigmoid and tanh functions respectively. The output $\mathbf{H}_v^0 \in \mathbb{R}^{n \times d_v \times T_3}$ represents temporally encoded features for modality $v \in \{M, L\}$.  
Stacking dilated convolution layers exponentially increases the receptive field, enabling efficient long-term modeling with reduced computational cost. We validate the efficiency of dilated convolutional operations in our experiments (see Subsection~\ref{Experimental_results}) by comparing their computational costs with those of VAR-based methods.

To explore the correlation of different factors from two modalities and then assess the contribution of each factor to causal structure learning, we compute a multi-factor similarity matrix for the $j^{\textrm{th}}$ system entity:
\begin{align}
\label{MDF_1_historical}
    \mathbf{C}_j^0 = \tanh(\mathbf{H}_M^0[j] \mathbf{W}^3 (\mathbf{H}_L^0[j])^\top),
\end{align}
where $\mathbf{W}^3 \in \mathbb{R}^{T_3 \times T_3}$ is a learnable projection.  
The similarity matrix $\mathbf{C}_j^0\in\mathbb{R}^{d_L \times d_M}$ captures inter-factor relationships between metric and log modalities for the historical data.  
We then compute attention-based importance weights to quantify each factor’s contribution to causal structure learning:
\begin{align}
    \nonumber \mathbf{Z}_L^0[j] &= \tanh(\mathbf{H}_L^0[j]\mathbf{W}^4 + \mathbf{H}_M^0[j]\mathbf{W}^5 \mathbf{C}_j^0), \\
    \mathbf{Z}_M^0[j] &= \tanh(\mathbf{H}_M^0[j]\mathbf{W}^5 + \mathbf{H}_L^0[j]\mathbf{W}^4 (\mathbf{C}_j^0)^\top), 
    \label{MDF_2_historical}\\
    \nonumber\mathbf{a}_L^0[j] &= \text{softmax}(\mathbf{w}^6 \mathbf{Z}_L^0[j]), \quad
    \mathbf{a}_M^0[j] = \text{softmax}(\mathbf{w}^7 \mathbf{Z}_M^0[j]),
\end{align}
where $\mathbf{W}^4, \mathbf{W}^5 \in \mathbb{R}^{T_3 \times T_3}$ and $\mathbf{w}^6, \mathbf{w}^7 \in \mathbb{R}^{T_4}$.  
The attention vectors $\mathbf{a}_v^0[j]$ measure the importance of each factor by encoding information from both modalities, capturing rich relationships for multi-modal and multi-dimensional data. With these attention vectors, the weighted modality representations are obtained as:
\begin{align}
\label{MDF_3_historical}
    \hat{\mathbf{H}}_v^0[j] = \sum_{k=1}^{d_v} \mathbf{a}_v^0[j,k] \cdot \mathbf{H}_v^0[j,k],
\end{align}
and then passed through a multi-layer perceptron (MLP) for factor recovery:
\begin{align}
\label{MDF_4_historical}
    \mathbf{O}_v^0 = \text{MLP}^0(\hat{\mathbf{H}}_v^0).
\end{align}
Overall, the multi-factor learning process (Eqs.~(\ref{MDF_1_historical}), (\ref{MDF_2_historical}), (\ref{MDF_3_historical}) and (\ref{MDF_4_historical})) is summarized as:
\begin{align}
\label{temp_mfl_historical}
    \hat{\mathbf{O}}_v^0 = MFL(\hat{\mathbf{H}}_v^0).
\end{align}

To learn causal relationships among entities, we apply a message-passing GNN (GraphSAGE~\citep{hamilton2017inductive}) to mimic fault propagation through a message-passing mechanism:
\begin{align}
\label{gcn_historical}
    \tilde{\mathbf{X}}_v^0 &= \sigma_2(\mathcal{A}_{\text{old}} (\hat{\mathbf{O}}_v^0 \oplus \mathbf{N}_v^0) \mathbf{W}^1), \\
    \nonumber \mathbf{N}_v^0[j] &= \frac{1}{|\mathcal{N}_j|} \sum_{k \in \mathcal{N}_j} \hat{\mathbf{O}}_v^0[k],
\end{align}
where $\mathbf{W}^1$ is the weight matrix, $\sigma_2$ denotes ReLU, $\oplus$ denotes concatenation, $\mathcal{A}_{\text{old}}$ is the historical causal graph, $\mathcal{N}_j$ represents node entity $j$'s neighbors,  and $\mathbf{N}_v^0$ aggregates neighbor information.  $\tilde{\mathbf{X}}_v^0$ predicts future values based on previous lagged data $\mathbf{\hat{X}}_v^0$, leveraging temporal dependencies captured by dilated convolutional neural networks. Finally, the training objective minimizes the forecasting error:
\begin{align}
\label{old_temporal_historical}
    \mathcal{L}_{t} = \frac{1}{n(d_L + d_M)} \sum_v \sum_{j=1}^n \sum_{k=1}^{d_v} 
    \mathbf{a}_v^0[j,k] \|\hat{\mathbf{X}}_v^0[j,k] - \tilde{\mathbf{X}}_v^0[j,k]\|^2.
\end{align}
The $\mathbf{a}_v^0[j,k]$ emphasize influential factors during temporal forecasting, reinforcing interpretable causal learning.

\paragraph{Dynamic Encoder.}
The \textit{Dynamic Encoder} extends the same architecture to model streaming data in the online setting, enabling adaptive causal updates. It aims to efficiently  model the long-term temporal dependencies and the causal relations among system entities and KPIs for the streaming data in the online setting. 
Given the current batch data $\hat{\mathbf{X}}_v^i$, temporal dependencies are encoded as:
\begin{align}
\label{temp_s1}
    \mathbf{H}_v^i &= \tanh(g(\hat{\mathbf{X}}_v^i, \mathbf{f}_3)) \odot \sigma(g(\hat{\mathbf{X}}_v^i, \mathbf{f}_4)), \\
    \hat{\mathbf{O}}_v^i &= MFL(\hat{\mathbf{H}}_v^i),
\end{align}
where $\mathbf{f}_3, \mathbf{f}_4$ are dilated convolutional kernels.  
We then propagate messages through the updated causal graph:
\begin{align}
\label{graphsage_batch}
    \tilde{\mathbf{X}}_v^i &= \sigma_2((\mathcal{A}_{\text{old}} + \Delta\mathcal{A}_v)(\hat{\mathbf{O}}_v^i \oplus \mathbf{N}_v^i) \mathbf{W}^2), \\
    \mathbf{N}_v^i[j] &= \frac{1}{|\mathcal{N}_j|} \sum_{k \in \mathcal{N}_j} \hat{\mathbf{O}}_v^i[k],
\end{align}
where $\Delta\mathcal{A}_v \in \mathbb{R}^{n \times n}$ is added to capture incremental causal changes in the current batch.  
Finally, we jointly minimize the reconstruction losses from historical and streaming data:
\begin{align}
\label{new_temporal}
    \nonumber\mathcal{L}_{t} &= \frac{1}{n(d_L + d_M)} \sum_v \sum_{j=1}^n \sum_{k=1}^{d_v}
    \mathbf{a}_v^i[j,k]\|\hat{\mathbf{X}}_v^i[j,k] - \tilde{\mathbf{X}}_v^i[j,k]\|^2
\end{align}
By integrating Eq.~(\ref{new_temporal}) with the message-passing mechanism in Eq.~(\ref{graphsage_batch}), the model continuously refines the causal adjacency matrix 
$\tilde{\mathcal{A}} = \mathcal{A}_{\text{old}} + \Delta\mathcal{A}_v$, capturing dynamic causal relations such as $\mathbf{X} \rightarrow \mathbf{y}$, where $\mathbf{X}$ denotes a potential root cause and $\mathbf{y}$ is the system KPI.

\subsection{Graph Fusion with Contrastive Multi-modal Learning}
To address the challenges of multi-modal learning (as discussed in Challenge~\textbf{C3} in Section~\ref{sec:intro}), we propose to enhance the relatedness between two modalities through \textbf{contrastive mutual information maximization}. Given the representations of historical data $\hat{\mathbf{H}}^0_v$ and current batch data $\hat{\mathbf{H}}^i_v$ extracted from both metric and log data, we maximize the mutual information between these modalities:
\begin{align}
    \mathcal{L}_{\mathrm{MI}} 
    &= \mathcal{I}_\phi(\hat{\mathbf{H}}_M^0, \hat{\mathbf{H}}_L^0) 
     + \mathcal{I}_\phi(\hat{\mathbf{H}}_M^i, \hat{\mathbf{H}}_L^i),
\end{align}
where $\mathcal{I}_\phi$ is the mutual information parameterized by a neural network $\phi$. Following the InfoNCE-style contrastive loss~\cite{oord2018representation, DBLP:conf/www/ZhengBWZH25, DBLP:journals/tmlr/ZhengFMH24, DBLP:conf/kdd/ZhengXZH22, DBLP:journals/corr/abs-2411-15623, DBLP:conf/sdm/ZhengZH23}, we approximate the mutual information by its lower bound:
\begin{align}
    \mathcal{I}_\phi(\hat{\mathbf{H}}_M^0, \hat{\mathbf{H}}_L^0)
    &:= \frac{1}{n} \sum_{j=1}^n 
    \log 
    \frac{
        \mathrm{sim}(\phi(\hat{\mathbf{H}}_M^0[j]), \phi(\hat{\mathbf{H}}_L^0[j]))
    }{
        \sum_k \mathrm{sim}(\phi(\hat{\mathbf{H}}_M^0[j]), \phi(\hat{\mathbf{H}}_L^0[k]))
    },
\end{align}
where $\mathrm{sim}(a,b)=\exp\!\left(\frac{a b^{\top}}{\|a\| \|b\|}\right)$ represents the exponential of cosine similarity between two entity representations $a$ and $b$.

To generate the causal graph for the current batch, directly summing the graphs from both modalities may yield overly dense or cyclic structures. This problem is amplified when one modality is of lower quality, as treating both modalities equally can obscure critical causal patterns. To mitigate this issue, we estimate the relative importance of the two modalities based on the correlation of multiple metrics. Using the similarity map for the current batch (\textit{i.e.}, $\mathbf{C}_j^i$), we compute the modality importance and fuse the two causal graphs as follows:
\begin{align}
    \nonumber 
    \mathbf{C}_j^i &= \tanh\!\big(\mathbf{H}_M^i[j] \mathbf{W}^8 (\mathbf{H}_L^i[j])^{\top}\big), \\
    \nonumber 
    s_M &= \frac{
        \sum_j \sum_l \exp\!\big(\sum_k \mathbf{C}_j^i[l, k]\big)
    }{
        \sum_j \Big[\sum_l \exp\!\big(\sum_k \mathbf{C}_j^i[l, k]\big)
        + \sum_k \exp\!\big(\sum_l \mathbf{C}_j^i[l, k]\big)\Big]
    }, \\
    \nonumber 
    \mathcal{A} &= (1 - s_M)\big(\mathcal{A}_{\mathrm{old}} + \Delta\mathcal{A}_L\big)
    + s_M\big(\mathcal{A}_{\mathrm{old}} + \Delta\mathcal{A}_M\big),
\end{align}
where $\mathbf{W}^8$ is a learnable weight matrix. This adaptive fusion mechanism ensures that higher-quality modalities contribute more to the causal graph update while maintaining structural sparsity and stability.

\noindent\textbf{Optimization.} The overall training objective is defined as:
\begin{align}
\label{ocean_overall}
    \nonumber \mathcal{L}_{\text{sparse}} & = \|\Delta\mathcal{A}_L\|_1 + \|\Delta\mathcal{A}_M\|_1, \\
    \mathcal{L} & = -\mathcal{L}_{MI} + \lambda_1 \mathcal{L}_{t} + \lambda_2 \mathcal{L}_{\text{sparse}} + \lambda_3 h(\mathcal{A}),
\end{align}
where $\|\cdot\|_1$ denotes the $\ell_1$-norm that enforces sparsity on the changes of the adjacency matrices. The sparsity regularization $\mathcal{L}_{\text{sparse}}$ encourages only a limited number of edge updates across modalities. The acyclicity constraint $h(\mathcal{A}) = \mathrm{tr}(e^{\mathcal{A} \odot \mathcal{A}}) - n = 0$ holds if and only if $\mathcal{A}$ is a directed acyclic graph~\citep{DBLP:conf/aistats/PamfilSDPGBA20}, where $\odot$ denotes the Hadamard product. $\lambda_1$, $\lambda_2$, and $\lambda_3$ are positive hyperparameters balancing the contribution of each component.

\noindent\textbf{Online Finetuning.} During the offline training stage, both the static and dynamic encoders are jointly optimized using Eq.~(\ref{ocean_overall}). In the online adaptation stage, the static encoder is frozen while the dynamic encoder is finetuned to efficiently adapt to streaming data with reduced computational overhead.

\subsection{Network Propagation-based Root Cause Identification} 
Malfunction effects often propagate from the root cause to its neighboring entities, implying that the immediate neighbors of anomalous KPIs may not necessarily be the true root causes. To accurately identify the root cause, we first derive a transition probability matrix based on the learned causal graph $\mathcal{G}$ and then employ a random walk with restart (RWR) algorithm~\citep{DBLP:conf/icdm/TongFP06} to simulate the propagation of malfunction signals:
\begin{equation}
    \mathbf{P}_{ij} = \frac{\mathcal{A}_{j,i}}{\sum_{k=1}^{n} \mathcal{A}_{k,i}},
\end{equation}
where $\mathbf{P}$ is the normalized adjacency matrix. During the random walk process, the model periodically restarts from the KPI node with probability $c \in [0, 1]$ to explore alternative propagation paths. The RWR update equation is formulated as:
\begin{equation}
    \mathbf{r}_{t+1} = (1-c)\mathbf{P}\mathbf{r}_{t} + c\mathbf{r}_{0},
\end{equation}
where $\mathbf{r}_{t}$ denotes the probability distribution over nodes at step $t$, and $\mathbf{r}_{0}$ is the initial distribution concentrated on the target KPI node. Upon convergence, the stationary distribution $\mathbf{r}_{t+1}$ reflects the influence score of each entity, and the top-$k$ entities with the highest scores are identified as the most probable root causes.

\noindent\textbf{Stopping Criterion.} 
As the online RCA process progresses across multiple data batches, both the inferred causal structure and the corresponding root cause rankings tend to stabilize. To avoid redundant computation, we introduce an adaptive stopping mechanism based on the \textit{rank-biased overlap} (RBO) metric~\citep{DBLP:journals/tois/WebberMZ10}, which quantifies the similarity between two ranked lists while emphasizing higher-ranked items. Given the root cause lists from consecutive batches, $R_{t-1}$ and $R_{t}$, their similarity is computed as:
\begin{align}
    \gamma &= \text{RBO}(R_{t-1}, R_{t}),
\end{align}
where $\gamma \in [0, 1]$. A larger $\gamma$ indicates a higher degree of stability in the root cause rankings. The online RCA process terminates automatically when the similarity score $\gamma$ exceeds a predefined threshold.

%% file: Experiment.tex
\section{Experiments}
In this section, we evaluate the effectiveness of our proposed \method\ by comparing it with state-of-the-art root cause analysis techniques. 
Additionally, we conduct a case study and an ablation study to further validate the assumptions outlined in the previous sections. 

\subsection{Experimental Setup}
\nop{\textbf{Datasets}. We evaluate the performance of our method, \method, using three public real-world datasets for root cause analysis: (1) \textbf{Product Review}\footnote{\url{https://github.com/KnowledgeDiscovery/rca_benchmark}}~\cite{zheng2024lemma}: This microservice system, dedicated to online product reviews, encompasses $234$ pods and is deployed across $6$ cloud servers. It recorded four system faults between May 2021 and December 2021. (2) \textbf{Online Boutique}~\cite{yu2023nezha}: This dataset represents a microservice system designed for e-commerce, and it includes five system faults. (3) \textbf{Train Ticket}~\cite{yu2023nezha}: This dataset is a microservice system for railway ticketing service with 5 system faults. All three datasets contain two modalities: system metrics and system logs.}

\begin{table*}[ht]
\caption{Results on Product Review dataset w.r.t different metrics.}
\centering
\begin{tabular}{*{10}{c}}
\hline      Modality      & Model         & PR@1      & PR@5      & PR@10     & MRR       & MAP@3     & MAP@5     & MAP@10 & Time (s)\\ \cline{1-9} \hline
\multirow{7}{*}{Metric Only} & PC           & 0         & 0           & 0      & 0.034           & 0        & 0       & 0  & 225.19\\   
                             & Dynotears    & 0         & 0.25         & 0.50      & 0.092     & 0	& 0.05	& 0.175  & 390.37\\   
                             & C-LSTM       & 0.25	   & 0.5	    & 0.5	  & 0.409	    & 0.417	& 0.45	& 0.475  & 1482.01\\   
                             & REASON       & 0.25	& \textbf{1.00}	    & \textbf{1.00}	      & 0.563	& 0.583	 & 0.75	 & 0.875 & 247.87\\
                             & CORAL        & 0.50	& \textbf{1.00}	    & \textbf{1.00}      & 0.750	& 0.833	 & 0.90	 & 0.950  & 146.46 \\
                             & Baro        & 0.50	& 0.50	    & 0.75      & 0.537	& 0.500	 & 0.50	 & 0.550 & 283.48\\
                             & $\epsilon$-Diagnosis         & 0	& 0	    & 0.25      & 0.038	& 0	 & 0	 & 0.050  & 367.13\\\hline  
\multirow{7}{*}{Log Only}    & PC           & 0     & 0     & 0        & 0.043	& 0  & 0   & 0 & 93.98\\   
                             & Dynotears    & 0	    & 0	    & 0.25	   & 0.058	 & 0  & 0	& 0.075  & 142.26\\   
                             & C-LSTM       & 0	   & 0	   & 0.25	  & 0.059	& 0  & 0   & 0.075  & 602.92\\   
                             & REASON       & 0     & 0     & 0.5	  & 0.088	& 0  & 0   & 0.100  & 129.17\\
                             & CORAL        & 0     & 0     & 0.5	  & 0.118	& 0  & 0   & 0.200  & 50.29 \\
                             & Baro        & 0.25	& 0.25  & 0.25  & 0.286	& 0.250	 & 0.25	 & 0.250 & 138.94\\
                             & $\epsilon$-Diagnosis         & 0	& 0	    & 0.25      & 0.039	& 0	 & 0	 & 0.050  & 149.23\\\hline   
\multirow{9}{*}{Multi-Modality}    & PC            & 0   & 0       & 0.25	& 0.054	 & 0	& 0	    & 0.075  & 300.26 \\  
                                & Dynotears     & 0   & 0.25	& 0.5	& 0.114	 & 0	& 0.05	& 0.225 & 426.78\\   
                                & C-LSTM        & 0.25	& 0.5	& 0.5	& 0.341	 & 0.250	& 0.35	& 0.425 & 1808.76 \\   
                                & REASON        & 0.5	& \textbf{1.00}    & \textbf{1.00}	    & 0.687  & 0.667  & 0.80	  & 0.900  & 303.5\\  
                                & MULAN        & 0.75	& \textbf{1.00}	   & \textbf{1.00}	    & 0.833	 & 0.833 & 0.90	 & 0.950  & 255.74\\
                                & CORAL        & 0.75	& \textbf{1.00}   & \textbf{1.00}	    & 0.875	 & 0.917	 & 0.95	 & 0.975  & 186.73 \\
                                & Baro        & 0.50	& 0.75	    & \textbf{1.00}      & 0.587	& 0.500	 & 0.60	 & 0.700 & 307.26\\
                                & $\epsilon$-Diagnosis         & 0	& 0	    & 0.25      & 0.042	& 0	 & 0	 & 0.075  & 402.33\\
                                & \method       & \textbf{1.00}   & \textbf{1.00}   & \textbf{1.00}     & \textbf{1.000}   & \textbf{1.000}  & \textbf{1.00}  & \textbf{1.000}  & \textbf{20.16}\\
\hline
\end{tabular}
\label{table_result_1}
\end{table*}

\begin{table*}[htp]
\caption{Results on Online Boutique w.r.t different metrics.}
\centering
\begin{tabular}{*{10}{c}}
\hline      Modality      & Model         & PR@1      & PR@3      & PR@5     & MRR       & MAP@2     & MAP@3     & MAP@5 & Time (s)\\ \cline{1-9} \hline
\multirow{8}{*}{Metric Only}  & PC  & 0.2  & 0.4  & 0.6  & 0.390  & 0.3  & 0.333  & 0.40 & 5.25\\
 & Dynotears	   & 0.2 & 0.4  & 0.4  & 0.344  & 0.2  & 0.267  & 0.32 & 14.56\\
 & C-LSTM       & 0  & 0.4  & 0.8  & 0.300  & 0.1  & 0.200  & 0.44 & 20.75\\
 & REASON       & 0.4  & 0.8  & \textbf{1.0}  & 0.617  & 0.5  & 0.600  & 0.76 & 3.23\\
 & CORAL        & 0.2  & \textbf{1.0}  & \textbf{1.0} & 0.600  & 0.6  & 0.733  & 0.84 & 2.99\\ 
 & Baro         & 0	& 0.8	    & \textbf{1.0}      & 0.417	& 0	 & 0.467	 & 0.68 & 3.46\\
 & $\epsilon$-Diagnosis         & 0	& 0.6	    & \textbf{1.0}      & 0.323	& 0	 & 0.267	 & 0.52  & 5.21\\\hline  
\multirow{8}{*}{Log Only}  & PC  & 0  & 0.4  & 0.6  & 0.257  & 0.1  & 0.200  & 0.32 & 3.88\\
 & Dynotears    & 0  & 0.2  & 0.6  & 0.207  & 0  & 0.067  & 0.24 & 10.23\\
 & C-LSTM       & 0  & 0.4  & 0.6  & 0.267  & 0.1  & 0.200  & 0.36 & 15.07\\
 & REASON       & 0.2  & 0.8  & 0.8  & 0.458  & 0.3  & 0.467  & 0.60 & 2.39\\
 & CORAL        & 0.2  & 0.6  & \textbf{1.0}  & 0.457  & 0.3  & 0.400  & 0.60 & 2.04\\
 &  Baro        & 0	& 0.6	    & 0.8      & 0.308	& 0	 & 0.267	 & 0.48 & 2.57\\
 & $\epsilon$-Diagnosis        & 0	& 0.4	    & 0.4      & 0.208	& 0	 & 0.133	 & 0.24  & 3.47\\\hline  
\multirow{10}{*}{Multi-Modality}  & PC  & 0.4  & 0.8  & \textbf{1.0}  & 0.573  & 0.4  & 0.533  & 0.68 & 6.78\\
 & Dynotears    & 0.2  & 0.6  & \textbf{1.0}  & 0.467  & 0.3  & 0.400  & 0.64 & 16.38\\
 & C-LSTM       & 0.2  & 0.4  & \textbf{1.0}  & 0.450  & 0.3  & 0.333  & 0.60 & 22.66\\
 & REASON       & 0.4  & \textbf{1.0}  & \textbf{1.0}  & 0.667  & 0.6  & 0.733  & 0.84  & 4.51\\
 & MULAN        & 0.4  & 0.8  & \textbf{1.0} & 0.617  & 0.5  & 0.600  & 0.76 & 4.96 \\
 & CORAL        & 0.4  & \textbf{1.0}  & \textbf{1.0}  & 0.700  & 0.7  & 0.800  & 0.88 & 3.63 \\
 & Baro         & 0.2	& \textbf{1.0}	    & \textbf{1.0}      & 0.567	& 0.2	 & 0.667	 & 0.80 & 4.88\\
 & $\epsilon$-Diagnosis         & 0	& 0.8	    & \textbf{1.0}      & 0.383	& 0	 & 0.400	 & 0.64  & 6.51\\
 & \method\  & \textbf{0.6}  & \textbf{1.0}  & \textbf{1.0}  & \textbf{0.800}  & \textbf{0.8}  & \textbf{0.867}  & \textbf{0.92} & \textbf{1.84}\\
\hline
\end{tabular}
\label{table_result_2}
\end{table*}

\textbf{Datasets}. We evaluate the performance of \method\ using three public real-world datasets: (1) \textbf{Product Review}\footnote{\url{https://lemma-rca.github.io/docs/data.html}}~\cite{zheng2024lemma}: A microservice system dedicated to online product reviews, encompassing $234$ pods deployed across $6$ cloud servers. It recorded four system faults between May 2021 and December 2021. (2) \textbf{Online Boutique}\footnote{\url{https://github.com/IntelligentDDS/Nezha}}~\cite{yu2023nezha}: A microservice system designed for e-commerce, including five system faults. (3) \textbf{Train Ticket}~\cite{yu2023nezha}: A microservice system for railway ticketing services, also with five system faults. All three datasets contain two modalities: system metrics and system logs.

\textbf{Evaluation Metrics}. We evaluate the effectiveness of the proposed model with three widely-used metrics~\cite{wang2023interdependent, DBLP:conf/iwqos/MengZSZHZJWP20}: (1) \textbf{Precision@K (PR@K)}: This metric measures the probability that the top-K predicted root causes are accurate. (2) \textbf{Mean Average Precision@K (MAP@K)}: It assesses the top-K predicted causes from an overall perspective. (3) \textbf{Mean Reciprocal Rank (MRR)}: This metric evaluates the ranking capability of the models. (4) \textbf{Time}: the training time (in seconds) for each batch of data. 

\textbf{Baselines}. We compare \method\ with eight methods: single modality RCA methods, including Dynotears~\cite{DBLP:conf/aistats/PamfilSDPGBA20},C-LSTM~\cite{DBLP:journals/pami/TankCFSF22}, REASON~\cite{wang2023interdependent}, multi-modality RCA method, i.e., MULAN~\cite{DBLP:conf/www/ZhengCHC24}, hypothesis based methods, PC~\cite{DBLP:journals/technometrics/Burr03}, $\epsilon$-Diagnosis~\cite{shan2019diagnosis}, BARO~\cite{pham2024baro}, one online RCA method (i.e., CORAL~\cite{wang2023incremental}). 

\textbf{Reproducibility.}
All experiments are conducted on a desktop running Ubuntu 18.04.5 with an Intel(R) Xeon(R) Silver 4110 CPU @2.10GHz and one 11GB GTX2080 GPU.  We use the Adam as the optimizer and we train the model for 100 iterations at each batch. We use two layers of dilated convolutional operations in the experiment. As for the stopping criteria, we terminate the identification process if the similarity $\gamma$ between the current batch and the previous batch is greater than 0.9 for three consecutive times. Each experiment is run with one trial. For Product Review dataset, we set the size of historical metric and log data to 8-hour intervals and each batch is set to be a 10-minute interval.
\textbf{Due to page limitation, we include the additional experiments, such as the experiment on Train Ticket dataset, the comparison with physical graph, time complexity analysis, three-modalities analysis, comparison with different architecture, and parameter analysis in a more complete version online\footnote{https://arxiv.org/abs/2410.10021}}.

\subsection{Performance Evaluation}
\paragraph{Experimental Results}
\label{Experimental_results}
In this subsection, we present the performance evaluation on Tables~\ref{table_result_1} and \ref{table_result_2} for various methods. Considering that many baseline methods (\eg, PC, C-LSTM, REASON, Dynotears, and Baro) are designed for the single-modality scenario, we assess their performance in both single-modality scenarios (\textit{e.g.}, system metrics only or system logs only) and the multi-modality case. We consider system logs as additional system metrics to enable the performance measurement of these single-modality RCA methods in the multi-modality scenario. We calculate an average ranking score based on the evaluation of different system metrics as the final result for all single-modality methods and \method. 
Our observations are as follows: (1) Compared to single-modality scenarios, most baseline methods benefit from leveraging multi-modality data across three distinct datasets. (2) As an online RCA method, CORAL outperforms all of the offline RCA methods with respect to seven metrics. (3). \method\ consistently outperforms all baseline methods across the three datasets.  
(4). The online RCA methods CORAL and \method\ have less training time compared with the offline RCA methods, while \method\ further reduces its computational cost to 1/9 compared with CORAL. The reduced computational cost is attributed to the efficiency of dilated convolutional operation and the design of the multi-factor attention mechanism. Notice that CORAL first learns the causal graph for each factor individually and then fuses the causal graphs, which is computationally expensive in the online setting.
Notably, \method\ exhibits a remarkable improvement in MRR on the Product Review dataset, excelling the second competitor (\textit{i.e.}, CORAL) by 12.5\%. Moreover, \method\ outperforms CORAL by 20\% and 10\% with respect to PR@1 and MAP@3, respectively. This is attributed to the assessment of multiple factors and the exploration of the correlation among different modalities.

\begin{figure}[ht]
\begin{center}
\vspace{-3mm}
\includegraphics[width=0.7\linewidth]{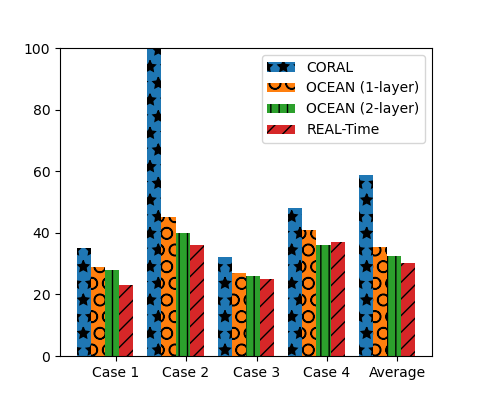}
\end{center}
\vspace{-5mm}
\caption{Identification time for four cases as well as the average identification time. }
\label{fig_identification_time}
\vspace{-4mm}
\end{figure}

\begin{table}[ht]
\caption{Ablation study on three datasets w.r.t MRR.}
\centering
\scalebox{0.9}{
\begin{tabular}{*{4}{c}}
\hline      Model       & Product Review     &  Online Boutique      &   Train Ticket   \\ \hline
\method             & \textbf{1.00}       & \textbf{0.8}                   & \textbf{0.381} \\
\method-F           & 0.75     & \textbf{0.8}                  & 0.331 \\ 
\method-M           & 0.875    & 0.7                   & 0.320 \\ 
\method-S           & 0.833    & 0.7                   & 0.345 \\
\hline
\end{tabular}}
\label{table_ablation_study}
\end{table}

\paragraph{Case Study}
\label{Case_study}
\nop{In this subsection, we aim to demonstrate the efficiency of the \method\ by evaluating the promptness of two online RCA methods on the Product Review dataset, \ie, CORAL and \method, shown in Figure~\ref{fig_identification_time}. Notice that we also evaluate the effectiveness of the long-term high-order Temporal causal structure learning module by varying the number of dilated convolutional layers, \eg., \method\ (1-layer) and \method\ (2-layer) for \method\ with 1-layer and 2-layer dilation convolutional networks, respectively. In Figure~\ref{fig_identification_time}, the y-axis is the batch index when an RCA method reaches the stopping criteria and the real-time suggests the actual system failure time for each case. A lower value of batch index implies the promptness of the online RCA method. Since CORAL fails to rank the real entity as the first possible root cause for system failure in case 2, we use the total number of batches to denote its detecting time for fair comparison.
By observation, we find (1) CORAL has around 10 epochs delay compared with the real-time, (2). \method\ (2-layer) has a lower detecting time in terms of batch index than \method\ (1-layer). This suggests that increasing the number of dilated convolutional networks indeed improves the promptness of real-time root cause identification, which verifies our assumption in Subsection~\ref{long_term_TCS} that stacking the dilated causal convolutional layers allows for modeling the longer temporal dependency.}
In this subsection, we evaluate the promptness of two online RCA methods on the Product Review dataset, CORAL and \method, as shown in Figure~\ref{fig_identification_time}. Note that we also evaluate the effectiveness of the long-term temporal causal structure learning module by varying the number of dilated convolutional layers in \method, specifically comparing configurations with one and two layers. In Figure~\ref{fig_identification_time}, the y-axis represents the batch index at which an RCA method meets the stopping criteria, and the real-time marker indicates the actual system failure time. A lower batch index value signifies faster identification of the ground-truth root cause by the RCA method. Notably, CORAL did not successfully rank the ground-truth root cause first in case 2, so we use the total number of batches to represent its detection time for a fair comparison. Our observations reveal that CORAL experiences about a 10-epoch delay relative to real-time in most cases, whereas \method\ (2-layer) achieves quicker detection than \method\ (1-layer). This improvement confirms our hypothesis that adding more dilated convolutional layers enhances the model's ability to capture longer temporal dependencies, as discussed in Subsection~\ref{long_term_TCS}.

\paragraph{Ablation Study}
In this subsection, we evaluate the effectiveness of individual components within the objective function of \method\ (Eq.~\ref{ocean_overall}). Specifically, we define \method-F and \method-M as variants that lack the multi-factor attention learning module and the contrastive multi-modal learning module, respectively, while \method-S removes the sparse constraint. The results, shown in Table~\ref{table_ablation_study}, indicate a significant performance degradation when any component is omitted. Specifically, removing the multi-factor attention module results in 25\% and 5\% performance drop on the Product Review dataset and Train Ticket dataset, respectively. Eliminating the contrastive multi-modal learning module leads to 12.5\% reduction on the Product Review dataset. These findings underscore the importance of each component in maintaining \method's high performance.

%% file: Conclusion.tex
\section{Conclusion}
\nop{In this paper, we investigates the challenging problem of online multi-modal root cause localization in microservice systems. We propose \method, a novel framework for online root causes localization by learning causal graphs from multi-modal data. \method\ leverages a tensor-based temporal causal structure learning module to capture long-term temporal dependency and causal relations among different node entities. To capture the relation of different factors and re-assess their contribution to causal graph learning, we introduce a multi-dimensional factor learning module with dilated convolution networks. Additionally, we present a contrastive mutual information maximization and graph fusion module that models the relationship between two modalities and fuses the causal graphs. We validated the effectiveness of \method\ through extensive experiments on three real-world datasets. }

In this paper, we investigate the challenging problem of online multi-modal root cause localization in microservice systems. We introduce \method, a novel online causal structure learning framework designed to effectively identify root causes using diverse data sources. \method\ utilizes a dilated convolutional neural network to capture long-term temporal dependencies and employs graph neural networks to establish causal relationships among system entities and key performance indicators. Additionally, we develop a multi-factor attention mechanism to evaluate and refine the contributions of various factors to the causal graph. Furthermore, \method\ incorporates a contrastive mutual information maximization-based graph fusion module to enhance interactions between different modalities and optimize their mutual information. The effectiveness of \method\ is validated through extensive experiments on three real-world datasets, demonstrating its robustness and efficiency.

%% file: Appendix.tex
\newpage

\appendix
\section{Appendix}


\begin{table*}[t]
\caption{Results on Train Ticket w.r.t different metrics.}
\centering
\begin{tabular}{*{10}{c}}
\hline      Modality      & Model         & PR@1      & PR@5      & PR@10     & MRR       & MAP@3     & MAP@5     & MAP@10  & Time (s) \\ \cline{1-9} \hline
\multirow{8}{*}{Metric Only} & PC  & 0  & 0  & 0.2  & 0.067  & 0  & 0  & 0.060  &  9.65\\
    & Dynotears  & 0  & 0  & 0  & 0.047  & 0  & 0  & 0 &  21.3\\
    & C-LSTM  & 0  & 0.2  & 0.2  & 0.097  & 0  & 0.08  & 0.140 & 30.63\\
    & REASON  & \textbf{0.2}  & \textbf{0.4}  & 0.6  & 0.323  & 0.2  & \textbf{0.36}  & 0.480 &  9.03\\
    & CORAL  & 0  & \textbf{0.4}  & \textbf{1.0}  & 0.184  & 0  & 0.16  & 0.500 & 5.72 \\
    & Baro        & 0	& 0.2	    & 0.8      & 0.145	& 0	 & 0.08	 & 0.243  & 9.68\\
    & $\epsilon$-Diagnosis        & 0	& 0.2	    & 0.4      & 0.105	& 0	 & 0.08	 & 0.143 & 11.32\\\hline 
\multirow{8}{*}{Log Only}     & PC  & 0  & 0.2  & 0.4  & 0.166 & 0.13  & 0.16  & 0.260 &  6.34\\
    & Dynotears  & 0  & 0  & 0.2  & 0.072 & 0  & 0  & 0.020 &  17.26\\
    & C-LSTM  & 0  & 0  & 0.2  & 0.072  & 0  & 0  & 0.020 & 21.40\\
    & REASON  & 0  & 0.2  & 0.6  & 0.126  & 0  & 0.08  & 0.280 & 7.58\\
    & CORAL  & 0  & 0.2  & 0.8  & 0.138  & 0  & 0.08  & 0.320 & 3.35\\
    & Baro       & 0	& \textbf{0.4}	    & 0.8      & 0.175	& 0 & 0.16	 & 0.314   & 7.89\\
    & $\epsilon$-Diagnosis       & 0	& 0.2	    & 0.4      & 0.106	& 0	 & 0.08	 & 0.114 & 9.77\\\hline 
\multirow{10}{*}{Multi-Modality}    &  PC  & 0  & 0  & 0.2  & 0.083  & 0  & 0  & 0.100 & 12.83\\
   & Dynotears  & 0  & \textbf{0.4}  & 0.6  & 0.141  & 0  & 0.16  & 0.320 & 27.82\\
   &  C-LSTM  & \textbf{0.2}  & \textbf{0.4}  & 0.6  & 0.294  & 0.2  & 0.28  & 0.360 & 36.76\\
   &  REASON  & \textbf{0.2}  & \textbf{0.4}  & 0.6  & 0.300  & 0.2  & 0.28  & 0.420 & 12.81\\
   &  MULAN  & \textbf{0.2}  & \textbf{0.4 } & \textbf{1.0}  & 0.317  & 0.2  & 0.28  & 0.460 &  11.42\\
   &  CORAL  & \textbf{0.2}  & \textbf{0.4 } & \textbf{1.0}  & 0.334  & 0.2  & 0.28  & 0.560 & 7.26\\
   & Baro        & 0 & \textbf{0.4}	    & \textbf{1.0}      & 0.254	& \textbf{0.33}	 & \textbf{0.36}	 & 0.414 & 13.74\\
   & $\epsilon$-Diagnosis         & 0	& 0	    & 0.6      & 0.114	& 0	 & 0	 & 0.143  & 16.28\\
   &  \method\  & \textbf{0.2}  & \textbf{0.4 } & \textbf{1.0}  & \textbf{0.381}  & \textbf{0.33}  & \textbf{0.36 } & \textbf{0.580} &  \textbf{3.22}\\
\hline
\end{tabular}
\label{table_result_3}
\end{table*}

\begin{table*}[t]
\caption{Results on Train Ticket w.r.t different metrics.}
\centering
\begin{tabular}{*{10}{c}}
\hline      Modality      & Model         & PR@1      & PR@5      & PR@10     & MRR       & MAP@3     & MAP@5     & MAP@10  & Time (s) \\ \cline{1-9} \hline
\multirow{8}{*}{Metric Only} & PC  & 0  & 0  & 0.2  & 0.067  & 0  & 0  & 0.060  &  9.65\\
    & Dynotears  & 0  & 0  & 0  & 0.047  & 0  & 0  & 0 &  21.3\\
    & C-LSTM  & 0  & 0.2  & 0.2  & 0.097  & 0  & 0.08  & 0.140 & 30.63\\
    & REASON  & \textbf{0.2}  & \textbf{0.4}  & 0.6  & 0.323  & 0.2  & \textbf{0.36}  & 0.480 &  9.03\\
    & CORAL  & 0  & \textbf{0.4}  & \textbf{1.0}  & 0.184  & 0  & 0.16  & 0.500 & 5.72 \\
    & Baro        & 0	& 0.2	    & 0.8      & 0.145	& 0	 & 0.08	 & 0.243  & 9.68\\
    & $\epsilon$-Diagnosis        & 0	& 0.2	    & 0.4      & 0.105	& 0	 & 0.08	 & 0.143 & 11.32\\\hline 
\multirow{8}{*}{Log Only}     & PC  & 0  & 0.2  & 0.4  & 0.166 & 0.13  & 0.16  & 0.260 &  6.34\\
    & Dynotears  & 0  & 0  & 0.2  & 0.072 & 0  & 0  & 0.020 &  17.26\\
    & C-LSTM  & 0  & 0  & 0.2  & 0.072  & 0  & 0  & 0.020 & 21.40\\
    & REASON  & 0  & 0.2  & 0.6  & 0.126  & 0  & 0.08  & 0.280 & 7.58\\
    & CORAL  & 0  & 0.2  & 0.8  & 0.138  & 0  & 0.08  & 0.320 & 3.35\\
    & Baro       & 0	& \textbf{0.4}	    & 0.8      & 0.175	& 0 & 0.16	 & 0.314   & 7.89\\
    & $\epsilon$-Diagnosis       & 0	& 0.2	    & 0.4      & 0.106	& 0	 & 0.08	 & 0.114 & 9.77\\\hline 
\multirow{10}{*}{Multi-Modality}    &  PC  & 0  & 0  & 0.2  & 0.083  & 0  & 0  & 0.100 & 12.83\\
   & Dynotears  & 0  & \textbf{0.4}  & 0.6  & 0.141  & 0  & 0.16  & 0.320 & 27.82\\
   &  C-LSTM  & \textbf{0.2}  & \textbf{0.4}  & 0.6  & 0.294  & 0.2  & 0.28  & 0.360 & 36.76\\
   &  REASON  & \textbf{0.2}  & \textbf{0.4}  & 0.6  & 0.300  & 0.2  & 0.28  & 0.420 & 12.81\\
   &  MULAN  & \textbf{0.2}  & \textbf{0.4 } & \textbf{1.0}  & 0.317  & 0.2  & 0.28  & 0.460 &  11.42\\
   &  CORAL  & \textbf{0.2}  & \textbf{0.4 } & \textbf{1.0}  & 0.334  & 0.2  & 0.28  & 0.560 & 7.26\\
   & Baro        & 0 & \textbf{0.4}	    & \textbf{1.0}      & 0.254	& \textbf{0.33}	 & \textbf{0.36}	 & 0.414 & 13.74\\
   & $\epsilon$-Diagnosis         & 0	& 0	    & 0.6      & 0.114	& 0	 & 0	 & 0.143  & 16.28\\
   &  \method\  & \textbf{0.2}  & \textbf{0.4 } & \textbf{1.0}  & \textbf{0.381}  & \textbf{0.33}  & \textbf{0.36 } & \textbf{0.580} &  \textbf{3.22}\\
\hline
\end{tabular}
\label{table_result_3}
\end{table*}

\label{Metrics}
\subsection{Evaluation Metrics}
We evaluate the model performance with the following four widely-used metrics~\citep{wang2023interdependent, DBLP:conf/iwqos/MengZSZHZJWP20}: \\
\noindent (1) \textbf{Precision@K (PR@K)}: It measures the probability that the top $K$ predicted root causes are relevant by: 
\begin{equation}
    \nonumber \text{PR@K} = \frac{1}{|\mathbb{A}|}\sum_{a \in \mathbb{A}}\frac{\sum_{i<k}R_a(i)\in V_a}{\min (K, |v_a|)}
\end{equation}
where $\mathbb{A}$ is the set of system faults, $a$ is one fault in $\mathbb{A}$, $V_a$ is the real root causes of $a$, $R_a$ is the predicted root causes of $a$, and i is the $i$-th predicted cause of $R_a$.\\
\noindent(2) \textbf{Mean Average Precision@K (MAP@K)}: It evaluates the top $K$ predicted causes from the overall perspective formulated as:
\begin{equation}
    \nonumber \text{MAP@K} = \frac{1}{K|\mathbb{A}|} \sum_{a \in \mathbb{A}} \sum_{i\leq j\leq K} PR@j
\end{equation}
where a higher value indicates a better performance.\\
\noindent(3) \textbf{Mean Reciprocal Rank (MRR)}: It assesses the ranking capability of models, defined as:
\begin{equation}
    \nonumber \text{MRR@K} = \frac{1}{|\mathbb{A}|}\sum_{a \in \mathbb{A}}\frac{1}{\text{rank}_{R_a}}
\end{equation}
where $\text{rank}_{R_a}$ is the rank number of the first correctly predicted root cause for system fault $a$.\\
\noindent(4) \textbf{Time}: Measures the training time (in seconds) for each batch of data. 

\subsection{Reproducibility}
All experiments are conducted on a desktop running Ubuntu 18.04.5 with an Intel(R) Xeon(R) Silver 4110 CPU @2.10GHz and one 11GB GTX2080 GPU. In the experiment, we set the size of historical metric and log data to 8-hour intervals and each batch is set to be a 10-minute interval. We use the Adam as the optimizer and we train the model for 100 iterations at each batch. We use two layers of dilated convolutional operations in the experiment. As for the stopping criteria, we terminate the identification process if the similarity $\gamma$ between the current batch and the previous batch is greater than 0.9 for three consecutive times. Each experiment is run with one trial.

\subsection{Time Complexity}
\label{time_complexity}
The time complexity of dilated convolution based causal structure learning is $O(nTkC)$. Here, $n$ is the number of system entities, $T$ is the sequence length, $L$ is the number of layers, $C$ is the number of filters, and $k$ is the filter size. The time complexity of multi-factor attention module is $O(d_Md_Ld)$, where $d$ is the hidden feature dimensionality, and $d_M$ and $d_L$ are the number of factors in two modalities. The time complexity of contrastive learning module is $O(n^2 d)$. Although Transformers (or LLMs) could capture temporal dependencies, they often come with high computational overhead. Specifically, the complexity of the Transformer is $O(nLT^2d)$.
Notice that $T$ can be very large on some large-scale datasets, such as Product Review dataset~\citep{zheng2024lemma}. As a result, Transformers are computationally more expensive than dilated convolutional neural networks. We further demonstrate the efficiency and effectiveness of our method compared to LSTM and Transformers in the ablation study presented in Table~\ref{table_ablation_study_2}.

\subsection{Baselines}
\label{baslines}
We compare \method\ with eight causal discovery based RCA methods: (1) \textbf{PC}~\citep{DBLP:journals/technometrics/Burr03}: A classic constraint-based algorithm that identifies the causal graph's skeleton using an independence test. (2) \textbf{Dynotears}~\citep{DBLP:conf/aistats/PamfilSDPGBA20}: A dynamic Bayesian network that uses vector autoregression model to capture the temporal dependency and learn the acyclic graph with DAG constraint. (3) \textbf{C-LSTM}~\citep{DBLP:journals/pami/TankCFSF22}: A root cause analysis method using LSTM to model temporal dependencies and capture nonlinear Granger causality. 
(4) 
\textbf{REASON}~\citep{wang2023interdependent}: A causal structure learning method that Learns both intra-level and inter-level causal relationships in interdependent networks. (5) \textbf{MULAN}~\citep{DBLP:conf/www/ZhengCHC24}: A multi-modal method that captures both modality-invariant and modality-specific representations. (6) \textbf{$\epsilon$-Diagnosis}~\citep{shan2019diagnosis}: A lightweight, unsupervised algorithm that detects root causes of long-tail latency in microservices by comparing time series similarity using $\epsilon$-statistics.
(7) \textbf{BARO}~\cite{pham2024baro}: An end-to-end RCA framework that combines change point detection and robust hypothesis testing to localize root causes in multivariate time-series data. (8) \textbf{CORAL}~\citep{wang2023incremental}: A VAR-based online RCA method that decouples state-invariant and state-dependent information. 

\subsection{Log Feature/Indicator Extraction}
\label{log_feature_extraction}
In this subsection, we provide the details of converting the raw data into the time series, though this is not within the scope of this work. Specifically, we first use the Drain parser~\cite{he2017drain} to transform the unstructured log event into structured log templates for each entity. Then, we partition the log data with fixed time windows, such as 5 minutes, and set time steps at 10 seconds. Within these intervals, we count the occurrence of each log template to derive the log frequency feature. The extraction of the log frequency feature is inspired by the insight that the recurrence of a log event template often correlates with its significance. For instance, when a microservice system experiences Distributed Denial of Service (DDoS) attacks, the system will produce an unusual volume of system logs, indicating abnormal activity. Thus, the log frequency provides the information to identify unusual patterns indicative of potential failure scenarios. In addition to log frequency, we also extract a second type of log feature known as the `golden signal.' Notice that different from log frequency, golden signal heavily relies on domain knowledge and it only focuses on the abnormal system logs. More specifically, we are only interested in some keywords, including `error,' `exception,' `critical,' `fatal', and various others indicative of system anomalies. By identifying these keywords within log event templates, we can discern abnormal occurrences for system failure localization. Similar to the frequency-based feature, we compute the number of abnormal log events to derive the golden signal-based feature.

\begin{table*}[h]
\caption{Ablation Study for different neural network architecture}
\centering
\begin{tabular}{*{9}{c}}
\hline \textbf{Product Review Dataset}	 & PR@1	 & PR@5	 & PR@10	 & MRR	 & MAP@3	 & MAP@5	 & MAP@10	 & Time (s) \\
\hline OCEAN	& 1	 		& 1		& 1		& 1		& 1		& 1		& 1		& 20.16 \\
\hline OCEAN-LSTM		& 0.25		& 1		& 1		& 0.542		& 0.583		& 0.75		& 0.875		& 546.37 \\
\hline OCEAN-Transformer		& 0.5		& 1		& 1		& 0.75		& 0.833		& 0.9		& 0.95		& 658.72 \\
\hline\hline \textbf{Train Ticket Dataset}	 & PR@1	 & PR@5	 & PR@10	 & MRR	 & MAP@3	 & MAP@5	 & MAP@10	 & Time (s) \\
\hline OCEAN		& 0.2		& 0.4		& 1.0		& 0.381		& 0.333		& 0.36		& 0.58		& 3.22 \\
\hline OCEAN-LSTM		& 0.2		& 0.6		& 0.8		& 0.342		& 0.2		& 0.36		& 0.54		& 12.6 \\
\hline OCEAN-Transformer		& 0.2		& 0.2		& 1		& 0.314		& 0.2		& 0.2		& 0.5		& 18.9 \\
\hline
\end{tabular}
\label{table_ablation_study_2}
\end{table*} 

\begin{table*}[ht]
\caption{Experimental results with more modality}
\centering
\begin{tabular}{*{9}{c}}
\hline \textbf{Product Review Dataset}	 & PR@1	 & PR@5	 & PR@10	 & MRR	 & MAP@3	 & MAP@5	 & MAP@10	 & Time (s) \\
\hline OCEAN			 &  1	 &  1	 &  1 	 &  1	 &  1	 &  1	 &  1	 &   20.1 \\
\hline OCEAN + trace		 &  1	 &  1	 &  1	 &  1	 &  1	 &  1	 &  1	 &   26.3  \\
\hline \textbf{Train Ticket}  &  PR@1  &   PR@5   &   PR@10  &   MRR  &   MAP@3   &  MAP@5  &  MAP@10   &  Time (s)  \\
\hline OCEAN			 &  0.2	 &  0.4	 &  1	 &  0.38	 &  0.33	 &  0.36	 &  0.58	 &   3.2 \\
\hline OCEAN + trace 		 &  0.2	 &  0.6	 &  1	 &  0.39	 &  0.33	 &  0.44	 &  0.62	 &   3.8 \\
\hline
\end{tabular}
\label{table_more_modality}
\end{table*} 

\subsection{Experimental Results with More Modalities}
\label{more_modalitieis}
OCEAN can naturally extend to include additional modalities, such as traces. These types of data can enhance the model's ability to capture complex interactions and dependencies within the system. We conducted additional experiments by incorporating traces into the AIOps and Train Ticket datasets. The results in Table~\ref{table_more_modality} demonstrated improved performance, as the inclusion of traces provided valuable context and enriched the causal structure learning. This additional information allows OCEAN to more accurately identify root causes and improve the precision of its analysis.

\begin{table}[h]
\caption{Comparison with physical graph}
\centering
\begin{tabular}{*{3}{c}}
\hline Graphs  &  SHD  &  AUROC \\
\hline Causal graph Learned From Metric Data         &  0.314  &  0.865   \\
\hline Causal graph Learned From Log Data             &  0.593  &  0.663  \\
\hline Fused Causal Graph  &  0.298  &  0.881 \\
\hline
\end{tabular}
\label{table_physical_graph}
\end{table}

\subsection{Comparison with Physical Graph}
\label{physical_graph}
Here, we evaluate the quality of the learned causal graph by comparing it with the physical dependency graph with two settings. In the first setting, we compared the causal graph learned by each modality (corresponding to the inter-modal graphs) and in the second setting, we compared the fused causal graph from two modality (corresponding to the intra-model graph). Following Dynotear~\cite{DBLP:conf/aistats/PamfilSDPGBA20}, we use AUROC and SHD as two metrics to quantify the difference between learned causal graphs and the physical dependency graph. The experimental results are shown in Table~\ref{table_physical_graph}.

\begin{figure*}[ht]
\begin{center}
\begin{tabular}{ccc}
\includegraphics[width=0.23\linewidth]{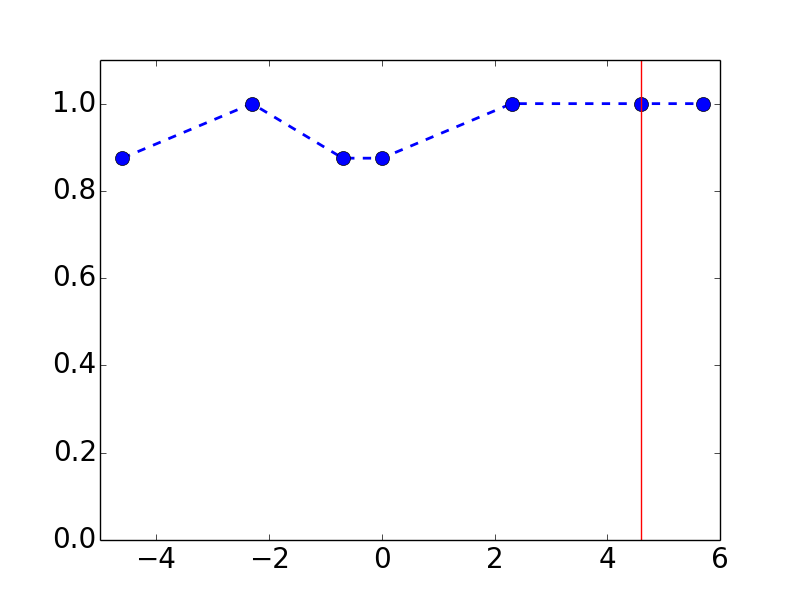} &
\includegraphics[width=0.23\linewidth]{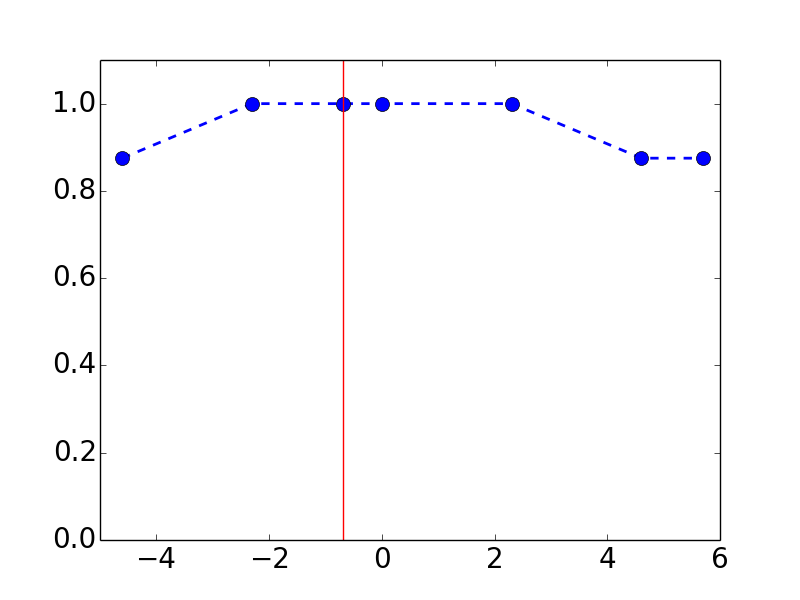} &
\includegraphics[width=0.23\linewidth]{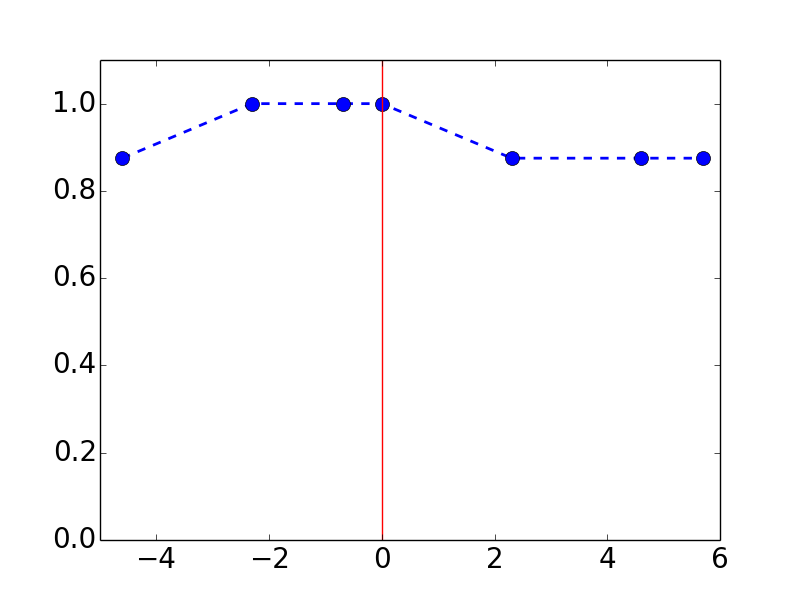} \\
(a) $\ln(\lambda_1)$ {\it w.r.t.} MRR&
(b) $\ln(\lambda_2)$ {\it w.r.t.} MRR &
(c) $\ln(\lambda_3)$ {\it w.r.t.} MRR
\end{tabular}
\end{center}
\caption{Parameter analysis on the Product Review dataset w.r.t MRR.}
\label{fig_parameter_analysis_1}
\end{figure*}

\begin{figure*}[htp]
\begin{center}
\begin{tabular}{ccc}
\includegraphics[width=0.23\linewidth]{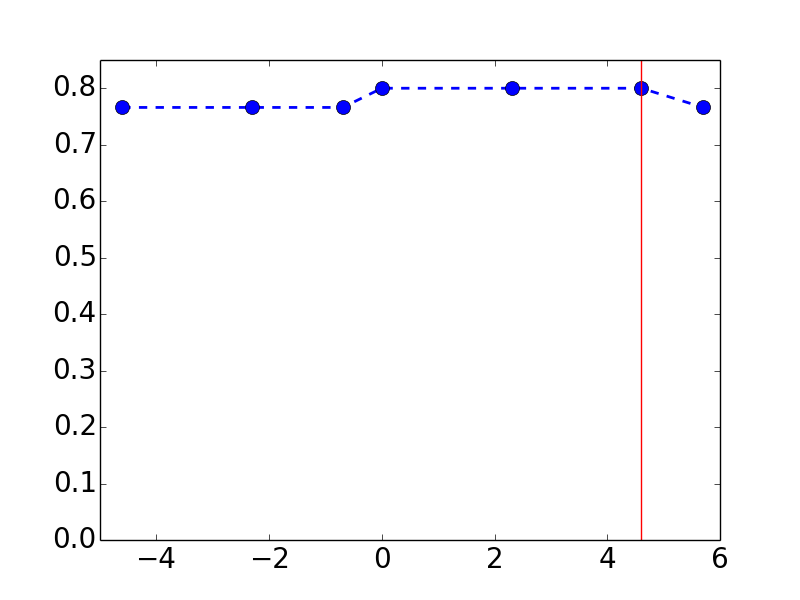} &
\includegraphics[width=0.23\linewidth]{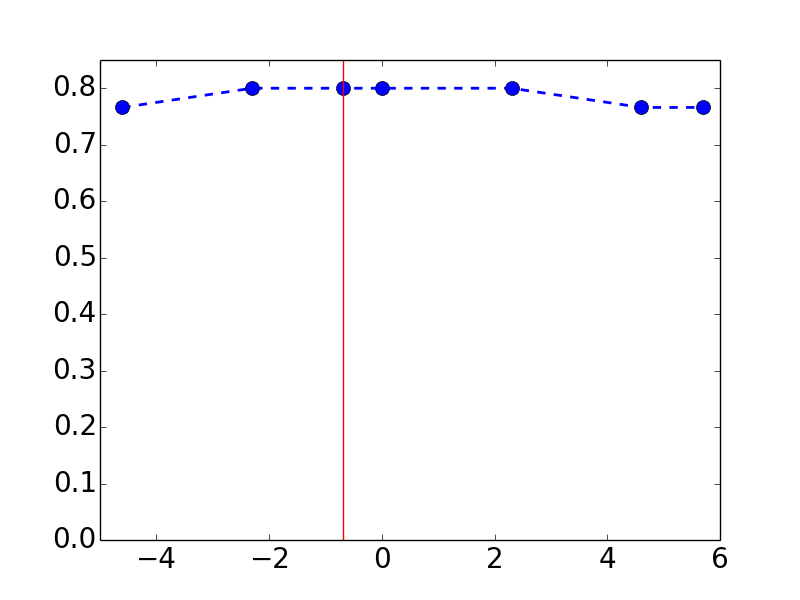} &
\includegraphics[width=0.23\linewidth]{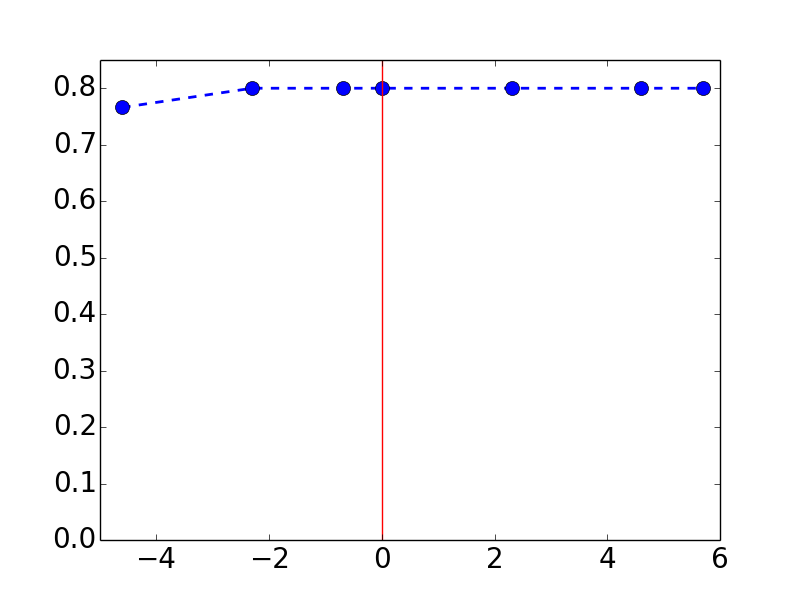} \\
(a) $\ln(\lambda_1)$ {\it w.r.t.} MRR&
(b) $\ln(\lambda_2)$ {\it w.r.t.} MRR &
(c) $\ln(\lambda_3)$ {\it w.r.t.} MRR
\end{tabular}
\end{center}
\caption{Parameter analysis on the Online Boutique dataset w.r.t MRR.}
\label{fig_parameter_analysis_2}
\end{figure*}

\begin{figure*}[htp]
\begin{center}
\begin{tabular}{ccc}
\includegraphics[width=0.23\linewidth]{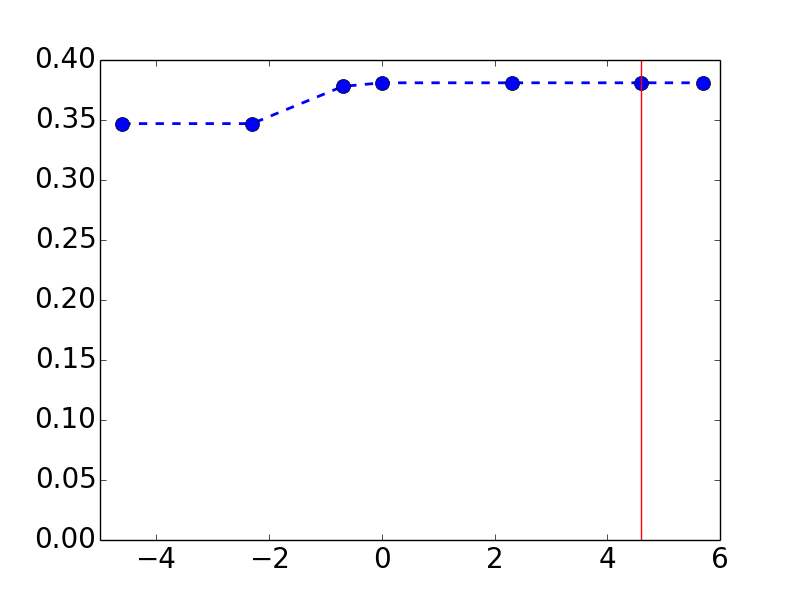} &
\includegraphics[width=0.23\linewidth]{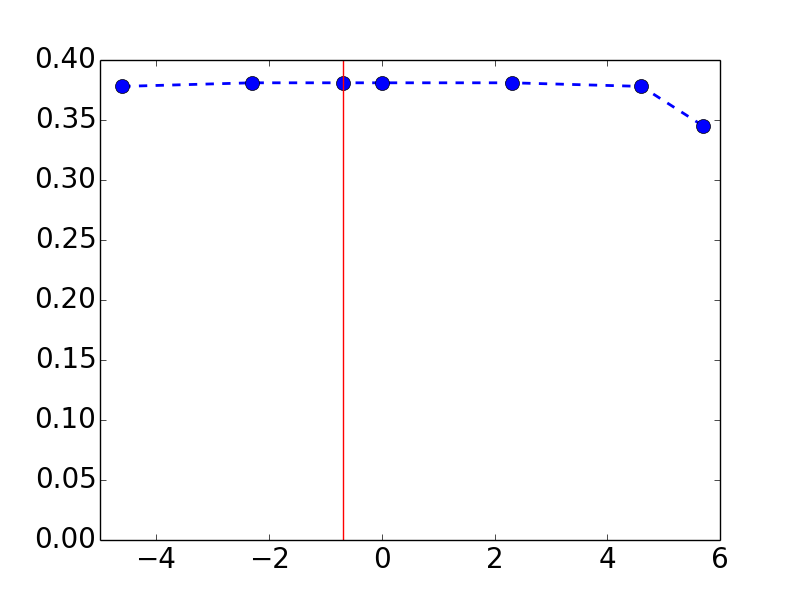} &
\includegraphics[width=0.23\linewidth]{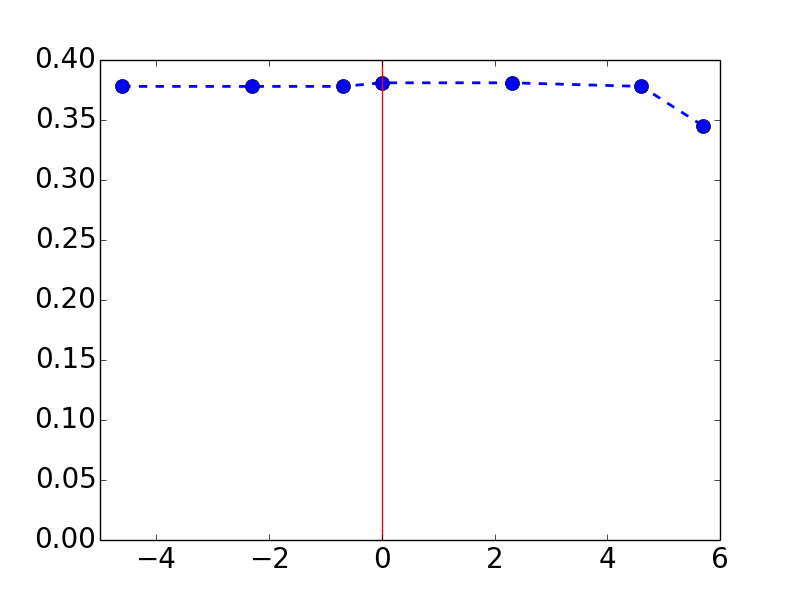} \\
(a) $\ln(\lambda_1)$ {\it w.r.t.} MRR&
(b) $\ln(\lambda_2)$ {\it w.r.t.} MRR &
(c) $\ln(\lambda_3)$ {\it w.r.t.} MRR
\end{tabular}
\end{center}
\caption{Parameter analysis on the Train Ticket dataset w.r.t MRR.}
\label{fig_parameter_analysis_3}
\end{figure*}

\subsection{Comparison with different architecture}
We evaluate the efficiency of our long-term temporal causal relationship module by replacing the dilated convolutional neural network with LSTM or Transformer model on the Product Review and Train Ticket datasets in Table~\ref{table_ablation_study_2}. The results showed a performance decline, demonstrating the effectiveness of our design.

\subsection{Parameter Analysis}
\label{parameter_analysis}
In this subsection, we present detailed parameter sensitivity analysis conducted on the Product Review, Online Boutique and Train Ticket datasets. we assess the impact of three parameters on the overall objective functions (Eq.~(\ref{ocean_overall})): $\lambda_1$, $\lambda_2$, and $\lambda_3$. The experimental results are displayed in Figures~\ref{fig_parameter_analysis_1}, \ref{fig_parameter_analysis_2} and \ref{fig_parameter_analysis_3}, showing the Mean Reciprocal Rank (MRR) on the Product Review dataset. On these figures, the x-axis represents $\ln(\lambda_i)$ for $i \in [1, 2, 3]$, and the y-axis shows the MRR score. Our analysis reveals that (1) We consistently observe that \method\ favors a larger value of $\lambda_1$ on these two datasets as the temporal causal structure learning module is crucial in capturing both temporal and causal dependency. (2) Conversely, $\lambda_2$ and $\lambda_3$ exhibit optimal performance at relatively lower values (\textit{e.g.}, $\ln({\lambda_2})=-0.7$ and $\ln(\lambda_3)=0$) on the Product Review dataset, with performance declining noticeably at higher levels. However, further reducing $\lambda_2$ and $\lambda_3$ also leads to diminished performance, which verifies the important role of sparse regularization and the acyclic constraint of the causal graph. Different from the parameter analysis on the Product Review dataset, we find that $\lambda_2$ and $\lambda_3$ are not very sensitive on the Online Boutique and Train Ticket datasets. We conjecture that this can be attributed to the small size of these two datasets and both sparse regularization and acyclic constraint contribute less to securing high performance on these two datasets than the Product Review dataset.

\subsection{Detailed Related Work}
\noindent\textbf{Root Cause Analysis}.
Current root cause analysis (RCA) methods can be categorized into two main branches: single-modal RCA methods and multi-modal RCA methods. Single-modal RCA methods primarily investigate causal relationships among system components using one type of data only~\citep{sporleder2019root,  duan2020root, DBLP:conf/iwqos/MengZSZHZJWP20, soldani2022anomaly, aggarwal2020localization, li2021practical}. For instance, Liu \textit{et al.}~\citep{DBLP:conf/icse/LiuH0LZGLOW21} generate a service call graph based on domain-specific software and rules, while Wang \textit{et al.}~\citep{wang2023interdependent} construct causal networks from time series data. However, these methods often exhibit suboptimal performance due to their reliance on single-modal data. To enhance accuracy, recent research integrates multi-modal data for RCA~\citep{yu2023nezha, DBLP:conf/ispa/HouJWLH21, DBLP:conf/www/ZhengCHC24}. Nezha~\citep{yu2023nezha} and PDiagnose~\citep{DBLP:conf/ispa/HouJWLH21} extract and combine information from each modality individually, while MULAN~\citep{DBLP:conf/www/ZhengCHC24} and MM-DAG~\citep{lan2023mm} consider interactions among modalities, constructing comprehensive causal graphs. Despite notable progress, these approaches are implemented offline, necessitating extensive data collection and retraining for new faults. Wang \textit{et al.}~\citep{wang2023incremental} enable online root cause identification by decoupling state-invariant and state-dependent information to learn a causal graph for root cause identification. However, their focus remains on single-modal data. Recently, large language model (LLM)-based approaches have emerged as a new research direction for learning causal relations in root cause identification, owing to the success of LLMs in tackling complex tasks~\citep{chen2024automatic, shan2024face, goel2024x, zhou2024causalkgpt, roy2024exploring, wang2023rcagent}. For example, Chen \textit{et al.}~\citep{chen2024automatic} introduce RCACopilot, an on-call system powered by LLMs to automate RCA for cloud incidents. Similarly, Shan \textit{et al.}~\citep{shan2024face} propose an approach that first identifies log messages indicating configuration-related errors, then localizes suspected root-cause configuration properties based on these log messages and LLM-generated configuration settings. While LLMs could effectively learn temporal dependencies within the metric data of individual system entities, they often struggle to capture interdependencies—such as causal relationships between different system entities—leading to higher computational costs. 
Unlike existing RCA methods, this paper addresses the online multi-modal RCA problem by uniquely modeling long-term temporal dependencies while simultaneously capturing the cross-modal correlation of multiple factors.

%% file: short_reference.bib
@inproceedings{DBLP:conf/nips/LuYBP16,
  author       = {Jiasen Lu and
                  Jianwei Yang and
                  Dhruv Batra and
                  Devi Parikh},
  title        = {Hierarchical Question-Image Co-Attention for Visual Question Answering},
  booktitle    = {NeurIPS 29},
  pages        = {289--297},
  year         = {2016}
}

@article{pham2024baro,
  title={Baro: Robust root cause analysis for microservices via multivariate bayesian online change point detection},
  author={Pham, Luan and Ha, Huong and Zhang, Hongyu},
  journal={Proceedings of the ACM on Software Engineering},
  volume={1},
  number={FSE},
  pages={2214--2237},
  year={2024},
  publisher={ACM New York, NY, USA}
}

@inproceedings{shan2019diagnosis,
  title={$\epsilon$-diagnosis: Unsupervised and real-time diagnosis of small-window long-tail latency in large-scale microservice platforms},
  author={Shan, Huasong and Chen, Yuan and Liu, Haifeng and Zhang, Yunpeng and Xiao, Xiao and He, Xiaofeng and Li, Min and Ding, Wei},
  booktitle={WWW},
  pages={3215--3222},
  year={2019}
}

@inproceedings{DBLP:conf/nips/VaswaniSPUJGKP17,
  author       = {Ashish Vaswani and
                  Noam Shazeer and
                  Niki Parmar and
                  Jakob Uszkoreit and
                  Llion Jones and
                  Aidan N. Gomez and
                  Lukasz Kaiser and
                  Illia Polosukhin},
  title        = {Attention is All you Need},
  booktitle    = {NeurIPS 30},
  pages        = {5998--6008},
  year         = {2017}
}

@inproceedings{DBLP:conf/ijcai/WuPLJZ19,
  author       = {Zonghan Wu and
                  Shirui Pan and
                  Guodong Long and
                  Jing Jiang and
                  Chengqi Zhang},
  title        = {Graph WaveNet for Deep Spatial-Temporal Graph Modeling},
  booktitle    = {IJCAI 2019},
  pages        = {1907--1913},
  publisher    = {ijcai.org},
  year         = {2019}
}

@inproceedings{DBLP:conf/aistats/PamfilSDPGBA20,
  author       = {Roxana Pamfil and
                  Nisara Sriwattanaworachai and
                  Shaan Desai and
                  Philip Pilgerstorfer and
                  Konstantinos Georgatzis and
                  Paul Beaumont and
                  Bryon Aragam},
  title        = {{DYNOTEARS:} Structure Learning from Time-Series Data},
  booktitle    = {The 23rd AISTATS},
  volume       = {108},
  pages        = {1595--1605},
  publisher    = {{PMLR}},
  year         = {2020}
}

@article{DBLP:journals/tois/WebberMZ10,
  author       = {William Webber and
                  Alistair Moffat and
                  Justin Zobel},
  title        = {A similarity measure for indefinite rankings},
  journal      = {{ACM} Trans. Inf. Syst.},
  volume       = {28},
  number       = {4},
  pages        = {20:1--20:38},
  year         = {2010}
}

@inproceedings{yu2023nezha,
  title={Nezha: Interpretable fine-grained root causes analysis for microservices on multi-modal observability data},
  author={Yu, Guangba and Chen, Pengfei and Li, Yufeng and Chen, Hongyang and Li, Xiaoyun and Zheng, Zibin},
  booktitle={the 31st FSE},
  pages={553--565},
  year={2023}
}

@inproceedings{lu2017log,
  author       = {Siyang Lu and
                  BingBing Rao and
                  Xiang Wei and
                  Byung{-}Chul Tak and
                  Long Wang and
                  Liqiang Wang},
  title        = {Log-based Abnormal Task Detection and Root Cause Analysis for Spark},
  booktitle    = {ICWS},
  pages        = {389--396},
  publisher    = {{IEEE}},
  year         = {2017}
}

@article{stock2001vector,
  title={Vector autoregressions},
  author={Stock, James H and Watson, Mark W},
  journal={Journal of Economic perspectives},
  volume={15},
  number={4},
  pages={101--115},
  year={2001},
  publisher={American Economic Association}
}

@article{lin2020limitations,
  title={Limitations of autoregressive models and their alternatives},
  author={Lin, Chu-Cheng and Jaech, Aaron and Li, Xin and Gormley, Matthew R and Eisner, Jason},
  journal={arXiv preprint arXiv:2010.11939},
  year={2020}
}

@article{oord2018representation,
  title={Representation learning with contrastive predictive coding},
  author={Oord, Aaron van den and Li, Yazhe and Vinyals, Oriol},
  journal={arXiv preprint arXiv:1807.03748},
  year={2018}
}

@inproceedings{DBLP:conf/iclr/KipfW17,
  author       = {Thomas N. Kipf and
                  Max Welling},
  title        = {Semi-Supervised Classification with Graph Convolutional Networks},
  booktitle    = {ICLR 2017},
  publisher    = {OpenReview.net},
  year         = {2017}
}

@inproceedings{DBLP:conf/icdm/TongFP06,
  author       = {Hanghang Tong and
                  Christos Faloutsos and
                  Jia{-}Yu Pan},
  title        = {Fast Random Walk with Restart and Its Applications},
  booktitle    = {Proceedings of ICDM 2006},
  pages        = {613--622},
  publisher    = {{IEEE} Computer Society},
  year         = {2006}
}

@article{DBLP:journals/pami/TankCFSF22,
  author       = {Alex Tank and
                  Ian Covert and
                  Nicholas J. Foti and
                  Ali Shojaie and
                  Emily B. Fox},
  title        = {Neural Granger Causality},
  journal      = {{IEEE} Trans. Pattern Anal. Mach. Intell.},
  volume       = {44},
  number       = {8},
  pages        = {4267--4279},
  year         = {2022}
}

@inproceedings{DBLP:conf/ispa/HouJWLH21,
  author       = {Chuanjia Hou and
                  Tong Jia and
                  Yifan Wu and
                  Ying Li and
                  Jing Han},
  title        = {Diagnosing Performance Issues in Microservices with Heterogeneous
                  Data Source},
  booktitle    = {ISPA},
  pages        = {493--500},
  publisher    = {{IEEE}},
  year         = {2021}
}

@article{DBLP:journals/technometrics/Burr03,
  author       = {Tom Burr},
  title        = {Causation, Prediction, and Search},
  journal      = {Technometrics},
  volume       = {45},
  number       = {3},
  pages        = {272--273},
  year         = {2003}
}

@inproceedings{DBLP:conf/iwqos/MengZSZHZJWP20,
  author       = {Yuan Meng and
                  Shenglin Zhang and
                  Yongqian Sun and
                  Ruru Zhang and
                  Zhilong Hu and
                  Yiyin Zhang and
                  Chenyang Jia and
                  Zhaogang Wang and
                  Dan Pei},
  title        = {Localizing Failure Root Causes in a Microservice through Causality
                  Inference},
  booktitle    = {28th IWQoS},
  pages        = {1--10},
  publisher    = {{IEEE}},
  year         = {2020}
}

@article{sporleder2019root,
  title={Root cause analysis on corrosive potential-induced degradation effects at the rear side of bifacial silicon PERC solar cells},
  author={Sporleder, Kai and Naumann, Volker and Bauer, Jan and Richter, Susanne and H{\"a}hnel, Angelika and Gro{\ss}er, Stephan and Turek, Marko and Hagendorf, Christian},
  journal={Solar Energy Materials and Solar Cells},
  volume={201},
  pages={110062},
  year={2019},
  publisher={Elsevier}
}

@article{duan2020root,
  title={Root cause analysis approach based on reverse cascading decomposition in QFD and fuzzy weight ARM for quality accidents},
  author={Duan, Panting and He, Zhenzhen and He, Yihai and Liu, Fengdi and Zhang, Anqi and Zhou, Di},
  journal={Computers \& Industrial Engineering},
  volume={147},
  pages={106643},
  year={2020},
  publisher={Elsevier}
}

@inproceedings{DBLP:conf/icse/LiuH0LZGLOW21,
  author       = {Dewei Liu and
                  Chuan He and
                  Xin Peng and
                  Fan Lin and
                  Chenxi Zhang and
                  Shengfang Gong and
                  Ziang Li and
                  Jiayu Ou and
                  Zheshun Wu},
  title        = {MicroHECL: High-Efficient Root Cause Localization in Large-Scale Microservice
                  Systems},
  booktitle    = {43rd ICSE },
  pages        = {338--347},
  publisher    = {{IEEE}},
  year         = {2021}
}

@inproceedings{DBLP:conf/kdd/0005LYNZSP22,
  author       = {Mingjie Li and
                  Zeyan Li and
                  Kanglin Yin and
                  Xiaohui Nie and
                  Wenchi Zhang and
                  Kaixin Sui and
                  Dan Pei},
  title        = {Causal Inference-Based Root Cause Analysis for Online Service Systems
                  with Intervention Recognition},
  booktitle    = {SIGKDD},
  pages        = {3230--3240},
  publisher    = {{ACM}},
  year         = {2022}
}

@inproceedings{DBLP:conf/nips/IkramCMSBK22,
  author       = {Azam Ikram and
                  Sarthak Chakraborty and
                  Subrata Mitra and
                  Shiv Kumar Saini and
                  Saurabh Bagchi and
                  Murat Kocaoglu},
  title        = {Root Cause Analysis of Failures in Microservices through Causal Discovery},
  booktitle    = {NeurIPS},
  year         = {2022}
}

@inproceedings{wang2023interdependent,
  author       = {Dongjie Wang and
                  Zhengzhang Chen and
                  Jingchao Ni and
                  Liang Tong and
                  Zheng Wang and
                  Yanjie Fu and
                  Haifeng Chen},
  title        = {Interdependent Causal Networks for Root Cause Localization},
  booktitle    = {Proceedings of the 29th {ACM} {SIGKDD} Conference on Knowledge Discovery
                  and Data Mining, {KDD} 2023, Long Beach, CA, USA, August 6-10, 2023},
  pages        = {5051--5060},
  publisher    = {{ACM}},
  year         = {2023}
}

@inproceedings{wang2023incremental,
  title={Incremental Causal Graph Learning for Online Root Cause Analysis},
  author={Wang, Dongjie and Chen, Zhengzhang and Fu, Yanjie and Liu, Yanchi and Chen, Haifeng},
  booktitle={SIGKDD},
  pages={2269--2278},
  year={2023}
}

@inproceedings{DBLP:conf/www/ZhengCHC24,
  author       = {Lecheng Zheng and
                  Zhengzhang Chen and
                  Jingrui He and
                  Haifeng Chen},
  title        = {{MULAN:} Multi-modal Causal Structure Learning and Root Cause Analysis
                  for Microservice Systems},
  booktitle    = {WWW 2024},
  pages        = {4107--4116},
  publisher    = {{ACM}},
  year         = {2024}
}

@inproceedings{lan2023mm,
  title={Mm-dag: Multi-task dag learning for multi-modal data-with application for traffic congestion analysis},
  author={Lan, Tian and Li, Ziyue and Li, Zhishuai and Bai, Lei and Li, Man and Tsung, Fugee and Ketter, Wolfgang and Zhao, Rui and Zhang, Chen},
  booktitle={SIGKDD},
  pages={1188--1199},
  year={2023}
}

@inproceedings{li2021practical,
  title={Practical root cause localization for microservice systems via trace analysis},
  author={Li, Zeyan and Chen, Junjie and Jiao, Rui and Zhao, Nengwen and Wang, Zhijun and Zhang, Shuwei and Wu, Yanjun and Jiang, Long and Yan, Leiqin and Wang, Zikai and others},
  booktitle={29th IWQOS},
  pages={1--10},
  year={2021},
  organization={IEEE}
}

@article{soldani2022anomaly,
    author={Soldani, Jacopo and Brogi, Antonio},
  title={Anomaly detection and failure root cause analysis in (micro) service-based cloud applications: A survey},
  journal={ACM Computing Surveys (CSUR)},
  volume={55},
  number={3},
  pages={1--39},
  year={2022},
  publisher={ACM New York, NY}
}

@article{podgorski2015measuring,
  title={Measuring operational performance of OSH management system--A demonstration of AHP-based selection of leading key performance indicators},
  author={Podg{\'o}rski, Daniel},
  journal={Safety science},
  volume={73},
  pages={146--166},
  year={2015},
  publisher={Elsevier}
}

@article{alanqary2021change,
  title={Change point detection via multivariate singular spectrum analysis},
  author={Alanqary, Arwa and Alomar, Abdullah and Shah, Devavrat},
  journal={NeurIPS},
  volume={34},
  pages={23218--23230},
  year={2021}
}

@inproceedings{bogner2017automatically,
  title={Automatically measuring the maintainability of service-and microservice-based systems: a literature review},
  author={Bogner, Justus and Wagner, Stefan and Zimmermann, Alfred},
  booktitle={IWSM Mensura},
  pages={107--115},
  year={2017}
}

@inproceedings{aggarwal2020localization,
  title={Localization of operational faults in cloud applications by mining causal dependencies in logs using golden signals},
  author={Aggarwal, Pooja and Gupta, Ajay and Mohapatra, Prateeti and Nagar, Seema and Mandal, Atri and Wang, Qing and Paradkar, Amit},
  booktitle={International Conference on Service-Oriented Computing},
  pages={137--149},
  year={2020},
  organization={Springer}
}

@inproceedings{he2017drain,
  title={Drain: An online log parsing approach with fixed depth tree},
  author={He, Pinjia and Zhu, Jieming and Zheng, Zibin and Lyu, Michael R},
  booktitle={ICWS},
  pages={33--40},
  year={2017},
  organization={IEEE}
}

@inproceedings{chen2024automatic,
  title={Automatic root cause analysis via large language models for cloud incidents},
  author={Chen, Yinfang and Xie, Huaibing and Ma, Minghua and Kang, Yu and Gao, Xin and Shi, Liu and Cao, Yunjie and Gao, Xuedong and Fan, Hao and Wen, Ming and others},
  booktitle={EuroSys},
  pages={674--688},
  year={2024}
}

@inproceedings{shan2024face,
  title={Face it yourselves: An llm-based two-stage strategy to localize configuration errors via logs},
  author={Shan, Shiwen and Huo, Yintong and Su, Yuxin and Li, Yichen and Li, Dan and Zheng, Zibin},
  booktitle={SIGSOFT},
  pages={13--25},
  year={2024}
}

@inproceedings{goel2024x,
  title={X-lifecycle Learning for Cloud Incident Management using LLMs},
  author={Goel, Drishti and Husain, Fiza and Singh, Aditya and Ghosh, Supriyo and Parayil, Anjaly and Bansal, Chetan and Zhang, Xuchao and Rajmohan, Saravan},
  booktitle={32nd ACM FSE},
  pages={417--428},
  year={2024}
}

@article{hamilton2017inductive,
  title={Inductive representation learning on large graphs},
  author={Hamilton, Will and Ying, Zhitao and Leskovec, Jure},
  journal={NeurIPS},
  volume={30},
  year={2017}
}

@article{zheng2024lemma,
  title={LEMMA-RCA: A Large Multi-modal Multi-domain Dataset for Root Cause Analysis},
  author={Zheng, Lecheng and Chen, Zhengzhang and Wang, Dongjie and Deng, Chengyuan and Matsuoka, Reon and Chen, Haifeng},
  journal={arXiv preprint arXiv:2406.05375},
  year={2024}
}

@article{zhou2024causalkgpt,
  title={CausalKGPT: industrial structure causal knowledge-enhanced large language model for cause analysis of quality problems in aerospace product manufacturing},
  author={Zhou, Bin and Li, Xinyu and Liu, Tianyuan and Xu, Kaizhou and Liu, Wei and Bao, Jinsong},
  journal={Advanced Engineering Informatics},
  volume={59},
  pages={102333},
  year={2024},
  publisher={Elsevier}
}

@inproceedings{roy2024exploring,
  title={Exploring llm-based agents for root cause analysis},
  author={Roy, Devjeet and Zhang, Xuchao and Bhave, Rashi and Bansal, Chetan and Las-Casas, Pedro and Fonseca, Rodrigo and Rajmohan, Saravan},
  booktitle={32nd ACM FSE},
  pages={208--219},
  year={2024}
}

@inproceedings{wang2023rcagent,
  title={Rcagent: Cloud root cause analysis by autonomous agents with tool-augmented large language models},
  author={Wang, Zefan and Liu, Zichuan and Zhang, Yingying and Zhong, Aoxiao and Fan, Lunting and Wu, Lingfei and Wen, Qingsong},
  booktitle={CIKM},
  year={2024}
}

@inproceedings{DBLP:conf/www/ZhouZZLH20,
  author       = {Dawei Zhou and
                  Lecheng Zheng and
                  Yada Zhu and
                  Jianbo Li and
                  Jingrui He},
  editor       = {Yennun Huang and
                  Irwin King and
                  Tie{-}Yan Liu and
                  Maarten van Steen},
  title        = {Domain Adaptive Multi-Modality Neural Attention Network for Financial
                  Forecasting},
  booktitle    = {{WWW} '20: The Web Conference 2020, Taipei, Taiwan, April 20-24, 2020},
  pages        = {2230--2240},
  publisher    = {{ACM} / {IW3C2}},
  year         = {2020}
}

@inproceedings{DBLP:conf/acl/0006ZJFJBH025,
  author       = {Zihao Li and
                  Lecheng Zheng and
                  Bowen Jin and
                  Dongqi Fu and
                  Baoyu Jing and
                  Yikun Ban and
                  Jingrui He and
                  Jiawei Han},
  editor       = {Wanxiang Che and
                  Joyce Nabende and
                  Ekaterina Shutova and
                  Mohammad Taher Pilehvar},
  title        = {Can Graph Neural Networks Learn Language with Extremely Weak Text
                  Supervision?},
  booktitle    = {Proceedings of the 63rd Annual Meeting of the Association for Computational
                  Linguistics (Volume 1: Long Papers), {ACL} 2025, Vienna, Austria,
                  July 27 - August 1, 2025},
  pages        = {11138--11165},
  publisher    = {Association for Computational Linguistics},
  year         = {2025}
}

@article{DBLP:journals/corr/abs-2502-08942,
  author       = {Zihao Li and
                  Xiao Lin and
                  Zhining Liu and
                  Jiaru Zou and
                  Ziwei Wu and
                  Lecheng Zheng and
                  Dongqi Fu and
                  Yada Zhu and
                  Hendrik F. Hamann and
                  Hanghang Tong and
                  Jingrui He},
  title        = {Language in the Flow of Time: Time-Series-Paired Texts Weaved into
                  a Unified Temporal Narrative},
  journal      = {CoRR},
  volume       = {abs/2502.08942},
  year         = {2025}
}

@article{DBLP:journals/corr/abs-2504-07394,
  author       = {Dongqi Fu and
                  Yada Zhu and
                  Zhining Liu and
                  Lecheng Zheng and
                  Xiao Lin and
                  Zihao Li and
                  Liri Fang and
                  Katherine Tieu and
                  Onkar Bhardwaj and
                  Kommy Weldemariam and
                  Hanghang Tong and
                  Hendrik F. Hamann and
                  Jingrui He},
  title        = {ClimateBench-M: {A} Multi-Modal Climate Data Benchmark with a Simple
                  Generative Method},
  journal      = {CoRR},
  volume       = {abs/2504.07394},
  year         = {2025}
}

@inproceedings{DBLP:conf/kdd/ZhengJLTH24,
  author       = {Lecheng Zheng and
                  Baoyu Jing and
                  Zihao Li and
                  Hanghang Tong and
                  Jingrui He},
  editor       = {Ricardo Baeza{-}Yates and
                  Francesco Bonchi},
  title        = {Heterogeneous Contrastive Learning for Foundation Models and Beyond},
  booktitle    = {Proceedings of the 30th {ACM} {SIGKDD} Conference on Knowledge Discovery
                  and Data Mining, {KDD} 2024, Barcelona, Spain, August 25-29, 2024},
  pages        = {6666--6676},
  publisher    = {{ACM}},
  year         = {2024}
}

@inproceedings{DBLP:conf/www/ZhengBWZH25,
  author       = {Lecheng Zheng and
                  John R. Birge and
                  Haiyue Wu and
                  Yifang Zhang and
                  Jingrui He},
  editor       = {Guodong Long and
                  Michale Blumestein and
                  Yi Chang and
                  Liane Lewin{-}Eytan and
                  Zi Helen Huang and
                  Elad Yom{-}Tov},
  title        = {Cluster Aware Graph Anomaly Detection},
  booktitle    = {Proceedings of the {ACM} on Web Conference 2025, {WWW} 2025, Sydney,
                  NSW, Australia, 28 April 2025- 2 May 2025},
  pages        = {1771--1782},
  publisher    = {{ACM}},
  year         = {2025}
}

@article{DBLP:journals/tmlr/ZhengFMH24,
  author       = {Lecheng Zheng and
                  Dongqi Fu and
                  Ross Maciejewski and
                  Jingrui He},
  title        = {DrGNN: Deep Residual Graph Neural Network with Contrastive Learning},
  journal      = {Trans. Mach. Learn. Res.},
  volume       = {2024},
  year         = {2024}
}

@inproceedings{DBLP:conf/kdd/ZhengXZH22,
  author       = {Lecheng Zheng and
                  Jinjun Xiong and
                  Yada Zhu and
                  Jingrui He},
  editor       = {Aidong Zhang and
                  Huzefa Rangwala},
  title        = {Contrastive Learning with Complex Heterogeneity},
  booktitle    = {{KDD} '22: The 28th {ACM} {SIGKDD} Conference on Knowledge Discovery
                  and Data Mining, Washington, DC, USA, August 14 - 18, 2022},
  pages        = {2594--2604},
  publisher    = {{ACM}},
  year         = {2022}
}

@article{DBLP:journals/corr/abs-2102-07751,
  author       = {Lecheng Zheng and
                  Yu Cheng and
                  Hongxia Yang and
                  Nan Cao and
                  Jingrui He},
  title        = {Deep Co-Attention Network for Multi-View Subspace Learning},
  journal      = {CoRR},
  volume       = {abs/2102.07751},
  year         = {2021}
}

@inproceedings{DBLP:conf/sdm/ZhengZH23,
  author       = {Lecheng Zheng and
                  Yada Zhu and
                  Jingrui He},
  editor       = {Shashi Shekhar and
                  Zhi{-}Hua Zhou and
                  Yao{-}Yi Chiang and
                  Gregor Stiglic},
  title        = {Fairness-aware Multi-view Clustering},
  booktitle    = {Proceedings of the 2023 {SIAM} International Conference on Data Mining,
                  {SDM} 2023, Minneapolis-St. Paul Twin Cities, MN, USA, April 27-29,
                  2023},
  pages        = {856--864},
  publisher    = {{SIAM}},
  year         = {2023}
}

@inproceedings{DBLP:conf/sdm/ZhengCH19,
  author       = {Lecheng Zheng and
                  Yu Cheng and
                  Jingrui He},
  editor       = {Tanya Y. Berger{-}Wolf and
                  Nitesh V. Chawla},
  title        = {Deep Multimodality Model for Multi-task Multi-view Learning},
  booktitle    = {Proceedings of the 2019 {SIAM} International Conference on Data Mining,
                  {SDM} 2019, Calgary, Alberta, Canada, May 2-4, 2019},
  pages        = {10--18},
  publisher    = {{SIAM}},
  year         = {2019}
}

@article{DBLP:journals/corr/abs-2411-15623,
  author       = {Mengfei Lan and
                  Lecheng Zheng and
                  Shufan Ming and
                  Halil Kilicoglu},
  title        = {Multi-label Sequential Sentence Classification via Large Language
                  Model},
  journal      = {CoRR},
  volume       = {abs/2411.15623},
  year         = {2024}
}
